\pdfoutput=1
\documentclass[11pt]{article}
\usepackage{amsmath}
\usepackage{amsmath}
\usepackage{amsfonts}

\usepackage[]{acl}

\usepackage{times}
\usepackage{latexsym}

\usepackage[T1]{fontenc}
\usepackage[utf8]{inputenc}

\usepackage{microtype}
\usepackage{array}
\usepackage{booktabs}
\usepackage{graphicx}
\usepackage{subfigure}
\usepackage{longtable}
\usepackage{multicol}
\usepackage{geometry}
\usepackage{tabularx}
\usepackage{xtab,afterpage}
\usepackage{caption}
\usepackage{times}
\usepackage{latexsym}
\usepackage{amsmath,amssymb,amsfonts}
\usepackage{multirow}
\usepackage{hyperref}
\usepackage{dblfloatfix}
\usepackage{verbatim}
\usepackage{colortbl}

\newcolumntype{L}[1]{>{\raggedright\let\newline\\\arraybackslash\hspace{0pt}}m{#1}}
\newcolumntype{C}[1]{>{\centering\let\newline\\\arraybackslash\hspace{0pt}}m{#1}}
\newcolumntype{R}[1]{>{\raggedleft\let\newline\\\arraybackslash\hspace{0pt}}m{#1}}

\usepackage{subfiles}
\usepackage{inconsolata}

\usepackage{graphicx}

\title{REGen: A Reliable Evaluation Framework for Generative Event Argument Extraction}
\author{Omar Sharif, Joseph Gatto, Madhusudan Basak,  Sarah M. Preum\\
Department of Computer Science, Dartmouth College \\
 \texttt{\{omar.sharif.gr, sarah.masud.preum\}@dartmouth.edu}}

\begin{document}
\maketitle

\begin{abstract}
% Event argument extraction identifies arguments for predefined event roles in text. Traditional evaluations rely on exact match (EM), requiring predicted arguments to match annotated spans exactly. However, this approach fails for generative models like large language models (LLMs), which produce diverse yet semantically accurate responses. EM underestimates performance by disregarding valid variations, implicit arguments (unstated but inferable), and scattered arguments (distributed across a document). To bridge this gap, we introduce \textbf{R}eliable \textbf{E}valuation framework for \textbf{Gen}erative event argument extraction (\textbf{REGen}), a framework that better aligns with human judgment. Our evaluation across six datasets shows that model performance improves by an average of 23.93 F1 points using REGen, which is lost under the EM approach. Human validation further confirms REGen’s effectiveness, achieving 87.67\% alignment with human assessments of argument correctness.

Event argument extraction identifies arguments for predefined event roles in text. Existing work evaluates this task with exact match (EM), where predicted arguments must align exactly with annotated spans. While suitable for span-based models, this approach falls short for large language models (LLMs), which often generate diverse yet semantically accurate arguments. EM severely underestimates performance by disregarding valid variations. Furthermore, EM evaluation fails to capture implicit arguments (unstated but inferable) and scattered arguments (distributed across a document). These limitations underscore the need for an evaluation framework that better captures models' actual performance.  

To bridge this gap, we introduce \textbf{REGen}, a \textbf{R}eliable \textbf{E}valuation framework for \textbf{Gen}erative event argument extraction. REGen combines the strengths of exact, relaxed, and LLM-based matching to better align with human judgment. Experiments on six datasets show that REGen reveals an average performance gain of +23.93 F1 over EM, reflecting capabilities overlooked by prior evaluation. Human validation further confirms REGen’s effectiveness, achieving 87.67\% alignment with human assessments of argument correctness.

\end{abstract}

\section{Introduction}
Information extraction is a key area in natural language processing \cite{gaizauskas-wilks-1998-information}. Event argument extraction (EAE) is a core information extraction task that transforms text into structured information. As EAE identifies and extracts event-specific arguments from texts, it is essential for a wide range of applications such as document understanding \cite{tong-etal-2022-docee}, misinformation detection \cite{wu-etal-2022-cross}, discourse understanding \cite{sharif-etal-2024-explicit}, pharmacovigilance \cite{sun-etal-2022-phee}. With the emergence of generative models (e.g., LLMs), EAE has gained significant attention in recent years \cite{zhang-etal-2024-ultra,zhang-etal-2025-survey}. However, previous studies \cite{gao2023exploringfeasibilitychatgptevent, sun-etal-2024-leveraging} indicate that LLMs perform poorly on EAE tasks. This is largely due to the disconnect between the nature of generative predictions and the exact span-based evaluation method commonly used for EAE \cite{huang-etal-2024-textee}.

\begin{figure}[t!]
  \centering
  \includegraphics[width =\linewidth]{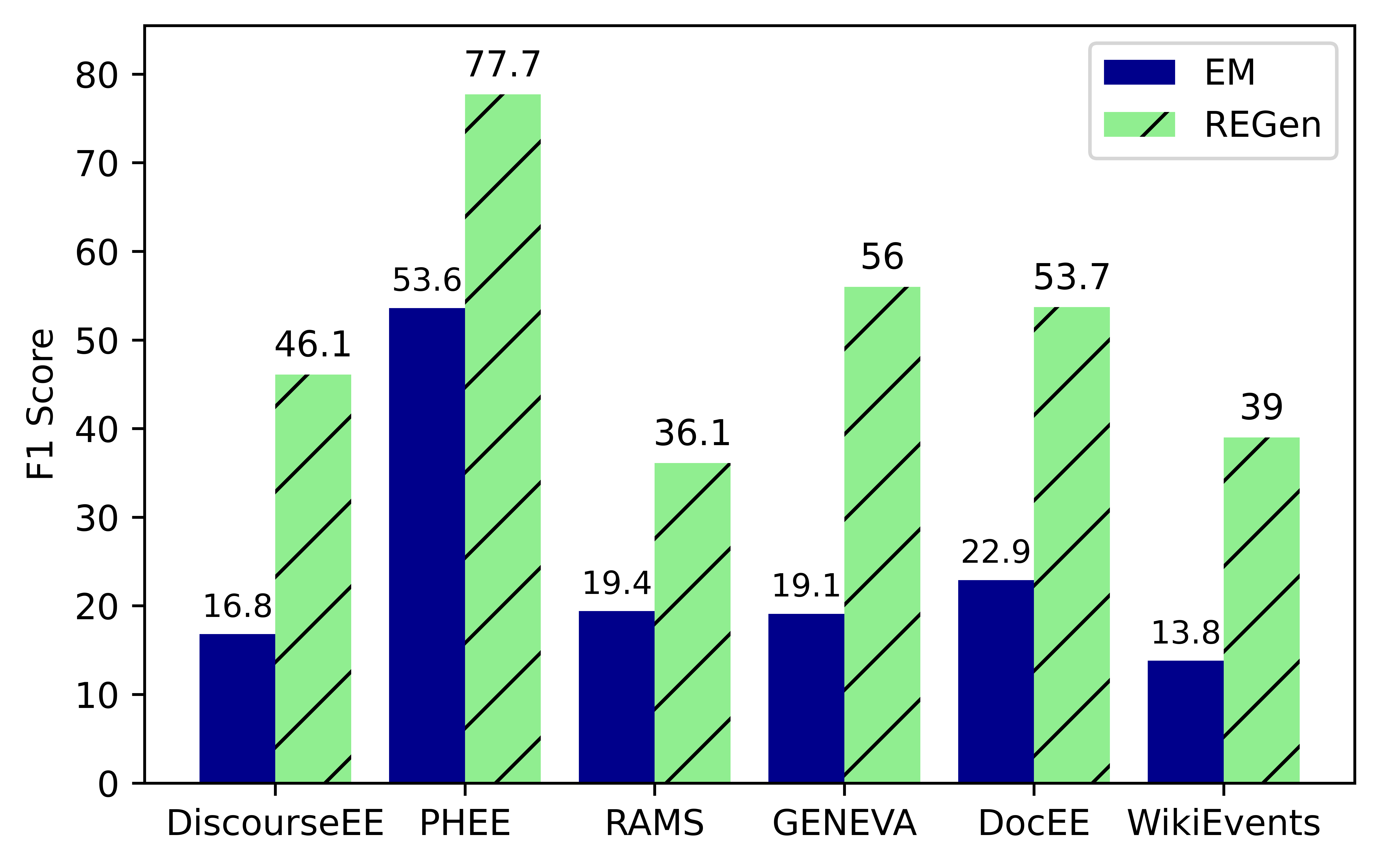}
 \caption{Performance comparison of the best-performing EAE model across six datasets under Exact Match (EM) and the REGen framework. The results highlight that, on average, EM underestimates model performance by 54.8\%, which is captured by REGen.}
 \label{intro-example-post}
\end{figure}

Span-based exact matching (EM) significantly underestimates the performance of LLMs as they often predict accurate arguments in surface forms that differ from the ground truth. For example, if the ground truth annotation for a role is \texttt{`pain relief'}, the model might output terms like \texttt{[alleviates pain, reducing discomfort, analgesia]}. Depending on the context, all or multiple of these outputs are correct, but none would be accepted by EM. Even minor variations would result in no match. Authors in \cite{sharif-etal-2024-explicit} highlighted that this problem is even more pronounced when evaluating the arguments composed of information from different parts of the text (scattered arguments) or the arguments that are not directly mentioned (implicit arguments).

\begin{figure*}[t!]
  \centering
  \includegraphics[width =0.94\linewidth]{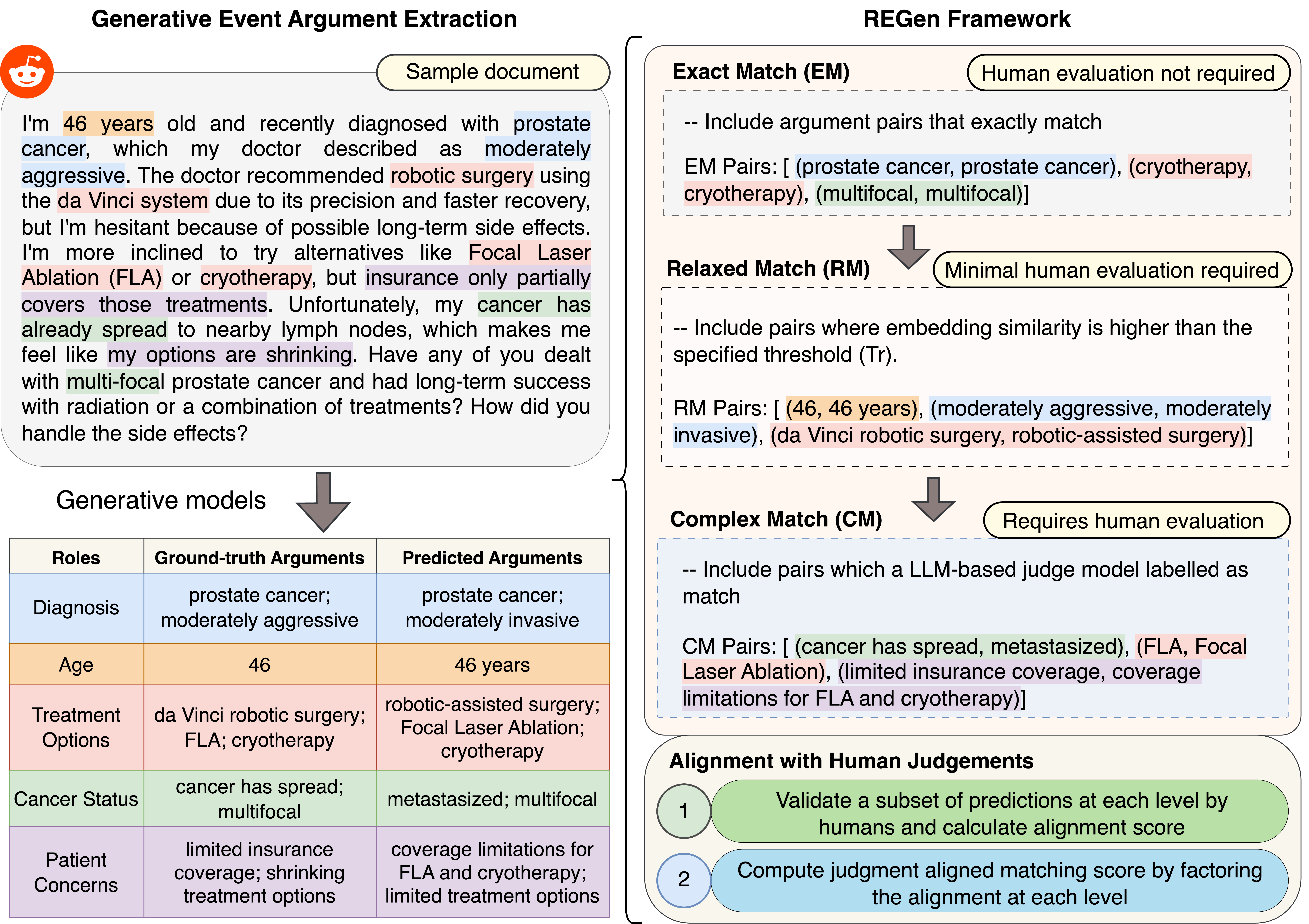}
 \caption{Proposed REGen evaluation framework for event argument extraction. \textbf{\textit{Left:}} An example of getting role-specific arguments from documents using generative models. Different colors indicate arguments for different roles. Semicolons separate multiple arguments for a role.  \textbf{\textit{Right:}} Illustration of the REGen's sequential evaluation process: Exact Match (\ref{subsection:exact-match}), Relaxed Match (\ref{subsection:relaxed-match}), and Complex Match (\ref{subsection:complex-match}) and Alignment with Human Judgments (\ref{subsection:judgement-alignment}). Only the arguments that do not match at the previous level are carried forward to the next level. Due to space constraints, the mathematical illustration of the framework is provided in the Appendix Figure \ref{eval-framework}.}
 \label{eval-framework-example}
\end{figure*}

Previous works have attempted to address these issues using embedding-based relaxed matching, which considers two arguments similar if they have high embedding similarity \cite{han2024empiricalstudyinformationextraction}. However, this approach fails to capture semantically similar arguments with different lexical forms and wrongly classifies arguments with high token overlap as similar \cite{sharif-etal-2024-explicit}. For example, in Figure \ref{eval-framework-example} for the role \textit{`patient concerns'}, the ground-truth argument \texttt{`limited insurance coverage'} and the predicted argument \texttt{`coverage limitations for FLA and cryotherapy'} refer to the same issue. Due to lexical variation, relaxed matching fails to capture this. In contrast, consider a role \textit{`date'} for which ground-truth and predicted arguments are \texttt{`18 April 2024'} and \texttt{`20 April 2024'}, respectively. These two arguments are different, but relaxed matching considers them the same due to high token overlap. Context is needed when evaluating these arguments. Recent work by  \citet{lu2024exactmatchsemanticallyreassessing} used LLMs as judges to identify similar arguments. This approach requires a large number of inferences, adding significant computational costs. Additionally, without human validation, LLM-based judgments can produce unreliable results.  Relying solely on relaxed match or judge-based approaches can overestimate performance by incorrectly classifying non-match arguments as matches, leading to inflated and unreliable model assessments. A detailed analysis of argument correctness by each method is shown in Table \ref{error-visualization}.

To address these limitations, we introduce \textbf{REGen}, a reliable evaluation framework for event argument extraction. REGen systematically combines the strengths of exact, relaxed, and LLM-based matching by \textbf{maximizing the evaluation reliability while minimizing the computation costs}. Figure \ref{eval-framework-example} illustrates the framework, and it is structured into four sequential phases: \textit{Exact Match (EM)}, \textit{Relaxed Match (RM)}, \textit{Complex Match (CM)}, and \textit{Alignment with Human Judgments}. 

The EM level filters arguments that match exactly, reducing computational costs for subsequent stages by eliminating obvious matches. This level does not require human evaluation as exact matches indicate perfect agreement with humans. The RM stage identifies arguments that are semantically similar, making evaluation robust to minor syntactic variations. This matching is performed based on the contextual embedding of the arguments. Setting up a high embedding similarity threshold ensures higher reliability and minimizes human evaluation. 

After filtering out exact and relaxed matches, unmatched arguments are carried forward for complex matching. The CM stage captures semantically similar arguments based on context despite lexical and/or syntactic differences. We leverage LLM as a judge \cite{NEURIPS2023_91f18a12} to determine argument similarity. Finally, in the judgment alignment stage, we propose a novel \textbf{Judgment Aligned Match (JAM)} score to factor in the scores from each level to account for misjudgments based on human validation. This framework ensures evaluation accuracy, cost-effectiveness, and better alignment with human judgments. 

To the best of our knowledge, this is the first systematic evaluation of LLMs on popular EAE datasets. Unlike prior studies \cite{lu2024exactmatchsemanticallyreassessing, huang-etal-2024-textee} that experimented on small test subsets sampled and merged from multiple datasets, we evaluate the complete test sets of the original datasets. This provides a more reliable assessment of LLMs' performances on these benchmarks and highlights their potential in solving the EAE task, which has been previously underestimated. Our key contributions are as follows.

\begin{itemize}
   
\item We present \textbf{REGen}, a \textbf{R}eliable \textbf{E}valuation framework for \textbf{Gen}erative event
argument extraction, minimizing inference costs and the need for human validation. REGen yields 87.67\% alignment with humans thus ensuring higher reliability. We also introduce a scoring mechanism to systematically measure how well REGen's evaluation aligns with human judgments. Finally, we curate a novel, human-annotated dataset with 900 samples to select LLM models as judges for EAE evaluation. % which we will release to enable future research in this direction.

\item We demonstrate the generalizability of REGen through extensive evaluation using multiple LLMs on six widely-used EAE datasets, including DiscourseEE \cite{sharif-etal-2024-explicit}, PHEE \cite{sun-etal-2022-phee}, RAMS \cite{ebner-etal-2020-multi}, GENEVA \cite{parekh-etal-2023-geneva}, DocEE \cite{tong-etal-2022-docee}, and WikiEvents \cite{li-etal-2021-document}. The results show an average improvement of 23.93 F1 points across all datasets while reducing inference costs by 41.2\% than the LLM-as-judge-only approach \cite{lu2024exactmatchsemanticallyreassessing}.

    % \item \textcolor{blue}{This seems a bit out-of-place: not sure if this is a separate contribution.}
    % To the best of our knowledge, this is the first systematic evaluation of generative models on popular EAE datasets. Unlike prior studies \cite{lu2024exactmatchsemanticallyreassessing, huang-etal-2024-textee} that experimented on small, combined test subsets, we evaluate the complete official test sets of the datasets. This provides a more accurate assessment of LLMs' performances on these benchmarks and highlights their potential in solving the EAE task, which has been previously underestimated.

\end{itemize}

\noindent
\textbf{Reproducibility:} Our code, evaluation framework,
the judge and alignment datasets, and other relevant resources are available at \href{https://github.com/Omar-Sharif/REGen}{https://github.com/Omar-Sharif/REGen}.

\section{REGen Framework}
\label{mlm-eval-framework}

% with corresponding ground-truth arguments
\subsection{Preliminaries}
\label{dataset-formatting}
\textbf{Document:} A document $D$ is a piece of text, which can be a sentence, a paragraph, or a full document.

\noindent
\textbf{Events, Roles and Arguments:} Events $(E)$ refer to occurrences or actions described in document $D$. A document can have multiple events. Each event is characterized by its roles $(R)$, which define the participants or entities involved. Arguments $(A)$ are specific details or attributes associated with these roles, providing context such as specific time, location, and other information. For example, consider the sentence: \textit{`Alice sent a package to Bob on Monday'}. The event here is \texttt{`Send'}, with potential roles such as sender, recipient, and time. The corresponding arguments for these roles are Alice, Bob, and Monday, respectively. 

\noindent 
\textbf{Generative Event Argument Extraction:}
This approach leverages generative models such as LLMs to extract arguments from the source document $D$. Given the source document, along with information about events and roles, the model generates a structured list of associated arguments.% $(A = [a_1, a_2, \ldots, a_{n}])$.
\subsection{Level-1: Exact Match}
\label{subsection:exact-match}
Let's assume we have a list of predicted and ground-truth argument strings for each role $R_i$ in a $D$. 
\[
P = [p_1, \ldots, p_{x1}], \quad
G = [g_1, \ldots, g_{y1}] 
\]
An exact match (EM) pair is defined when $p_i = g_j$, forming a list of EM pairs such as \( [(p_1, g_2), (p_3, g_5), \ldots, (p_x, g_y)] \). Precision and recall for the EM level are computed as:

{
\small
\[
\text{EM}_p = \frac{\text{NP}_e}{|P|}, \quad
\text{EM}_r = \frac{\text{NG}_e}{|G|}
\]
}
% \[
% \text{EM}_{f1} = \frac{2 * \text{EM}_{p} * \text{EM}_{r}}{\text{EM}_{p}+\text{EM}_{r}}
% \]
Here, \(\text{NP}_e\) and \(\text{NG}_e\) represent the number of correctly predicted arguments from the predicted (\(P\)) and ground-truth (\(G\)) argument lists, respectively. Note that \(\text{NP}_e = \text{NG}_e\) under exact match. 

\subsection{Level 2: Relaxed Match}
\label{subsection:relaxed-match}
Predicted and ground-truth argument lists are updated by removing arguments matched in Level-1. 
\[
P_{rm} = [p_1, \ldots, p_{x2}], \quad
G_{rm} = [g_1, \ldots, g_{y2}] 
\]
%$[(p'_1, g'_1), \dots (p'_{l_{x2}}, g'_{l_{y2}})]$
We compute the embedding-based similarity for all possible argument pairs of $P_{rm}$ and $G_{rm}$. A pair is considered a relaxed match (RM) if its similarity score exceeds the predefined threshold \(T_r\). The threshold selection method is described in section \ref{threshold-selection-process}.  The resulting list of RM pairs is a subset of all possible pairs. Precision and recall for relaxed matching are computed as:

{\small
\[
\text{RM}_{p} = \frac{\text{NP}_e + \text{NP}_r}{|P|}, \quad
\text{RM}_{r} = \frac{\text{NG}_e + \text{NG}_r}{|G|}
\]
}
% \[
% \text{RM}_{f1} = \frac{2 * \text{RM}_{p} * \text{RM}_{r}}{\text{RM}_{p}+\text{RM}_{r}}
% \]
Here, \(\text{NP}_r\) and \(\text{NG}_r\) represent the arguments matched under relaxed conditions form $P_{rm}$ and $G_{rm}$, while \(\text{NP}_e\) and \(\text{NA}_e\) are taken from Level 1. Relaxed matching allows an argument (\(p_x\) or \(g_y\)) to appear in multiple pairs where the similarity exceeds \(T_r\). To avoid overcounting, separate counts ($\text{NP}_r$, $\text{NG}_r$) are maintained for the number of arguments correctly matched from the prediction list and the ground-truth list.

\subsection{Level 3: Complex Match}
\label{subsection:complex-match}
After exact and relaxed matching, unmatched arguments are carried forward for complex matching. 
\[
P_{cm} = [p_1, \ldots, p_{x3}], \quad
G_{cm} = [g_1, \ldots, g_{y3}] 
\]
For the possible pairs from these lists, a preselected judge model determines similarity based on context. Details on how a judge model is selected for complex matching are discussed in section \ref{section: judge-selection}. If a pair is predicted as similar, it is added to the complex match (CM) pair list. Precision and recall for complex matching are computed as:

{\small
\[
\text{CM}_{p} = \frac{\text{NP}_e + \text{NP}_r + \text{NP}_c}{|P|}, \quad
\text{CM}_{r} = \frac{\text{NG}_e + \text{NG}_r + \text{NG}_c}{|G|}
\]
}
% \[
% \text{CM}_{f1} = \frac{2 * \text{CM}_{p} * \text{CM}_{r}}{\text{CM}_{p}+\text{CM}_{r}}
% \]
Here, \(\text{NP}_c\) and \(\text{NG}_c\) represent arguments correctly matched by the judge model. Similar to relaxed match, separate counts ensure that arguments are not overcounted when they appear in multiple matches. \(\text{NP}_e\), \(\text{NG}_e\), \(\text{NP}_r\), and \(\text{NG}_r\) are precomputed values from previous levels. 

\textbf{Generic Equation:}
We define three match levels \(L = [\text{EM}, \text{RM}, \text{CM}]\). \(\text{NP}_{e}\), \(\text{NP}_{r}\), and \(\text{NP}_{c}\) denote the number of correctly predicted arguments from the prediction list (\(P\)), while \(\text{NG}_{e}\), \(\text{NG}_{r}\), and \(\text{NG}_{c}\) represent the number of correctly matched arguments from the ground-truth list (\(G\)) at each level. Finally, precision, recall, and F1-score for a given level are calculated using the Equations \ref{eq: precision-recall}-\ref{eq: f1}.

{
\small
\begin{equation}
   \text{P}_l = \frac{\sum_{i=1}^{l} \text{NP}_{i}}{|P|},  \quad
   \text{R}_l = \frac{\sum_{i=1}^{l} \text{NG}_{i}}{|G|}
   \label{eq: precision-recall}
\end{equation}
\begin{equation}
   \text{F1}_l = \frac{2 * \text{P}_l * \text{R}_l}{\text{P}_l + \text{R}_l}
   \label{eq: f1}
\end{equation}
}
\subsection{Alignment with Human Judgments}
\label{subsection:judgement-alignment}

In Levels 2 (Relaxed Match) and 3 (Complex Match), the performance can be overestimated if the relaxed matching model or the complex match judge incorrectly classifies a non-match pair as a match. To account for this overestimation, we introduced a novel \textbf{Judgment Aligned Match (JAM)} Score, which penalizes the counts on each level based on the deviation from human judgment. %This adjustment offers a more reliable estimate of the model’s performance when we use LLMs as a judge model.

We first calculate the \textit{deviation rate} of a matching model $(M)$ on a dataset $(DT)$ by measuring the number of disagreements between the model and the human evaluator. The deviation rate is computed using equation \ref{eq: deviation-rate}. Addition details on the deviation rate or the alignment calculation are provided in Appendix \ref{appendix:alignment-calculation}.

\begin{equation}
    \text{E}_{(M, DT)} = \frac{\text{N}_d}{\text{N}_o}
    \label{eq: deviation-rate}
\end{equation}

Here, $\text{N}_d$ and $\text{N}_o$ denote the number of disagreements and the total number of observations, respectively. We calculate the \textbf{JAM Score} for a dataset factoring the model’s score at each matching level (EM, RM, CM) by the deviation rate of that level following equations \ref{eq: jam-precision}-\ref{eq: jam-f1}.

{\small
\begin{equation}
   \text{JAM}_p = \frac{\sum_{i=1}^{L} ((1-\text{E}_{i}) * \text{NP}_{i})}{|P|}
      \label{eq: jam-precision}
\end{equation}

\begin{equation}
   \text{JAM}_r = \frac{\sum_{i=1}^{L} ((1-\text{E}_{i}) * \text{NG}_{i})}{|G|}
   \label{eq: jam-recall}
\end{equation}

\begin{equation}
   \text{JAM}_{f1} = \frac{2 * \text{JAM}_p * \text{JAM}_r}{\text{JAM}_p + \text{JAM}_r}
   \label{eq: jam-f1}
\end{equation}
}

The JAM Score improves the alignment with human judgment, providing a more reliable reflection of the model’s true performance. 

\section{REGen Implementation Details}
\subsection{Threshold Selection for Relaxed Match}
\label{threshold-selection-process}
Usually, the model-predicted arguments contain the core words from the corresponding ground-truth arguments \cite{sharif-etal-2024-explicit, lu2024exactmatchsemanticallyreassessing}. While these predictions may have redundant words or miss some surrounding words, such discrepancies do not alter the overall semantics. We can identify these variations by using a high threshold relaxed match for accurate evaluation. We consider two arguments similar if their semantic similarity score exceeds 0.85, calculated using SBERT embeddings \cite{reimers-gurevych-2019-sentence}. 

This threshold is determined as follows. We tested three thresholds: 0.95, 0.85, and 0.75, across 500 argument pairs sourced from the six EAE datasets evaluated. The disagreement (error) rates were 0.0\%, 1.78\%, and 8.33\% for these thresholds, respectively. Although the 0.95 threshold yielded perfect agreement with human assessments, it allowed us to filter only a limited number of arguments. Conversely, the 0.75 threshold led to many incorrect matches. Therefore, we selected 0.85 as the optimal threshold. Our judgment alignment step ensures that our results are reliable and not inflated due to misjudgments.
 
\subsection{Judge Selection for Complex Match}
\label{section: judge-selection}
Studies show that LLMs achieve a strong correlation with human judgment across various tasks \cite{fu-etal-2024-gptscore,liu-etal-2023-g}. We also used LLMs to determine whether the ground truth and predicted arguments match. This approach makes the evaluation scalable across datasets and models. 

\textbf{Judge data annotation:} We construct a \textit{judge dataset comprising 900 argument pairs} (150 pairs per dataset) to select the
best judge model. Specifically, we randomly select pairs not matched under exact or relaxed criteria, meaning they inherently represent challenging or ambiguous cases. Each
pair is annotated as \textit{`match'} or \textit{`non-match'} by a human annotator. A second human verifies the labels,
and disagreements are resolved through discussion
to finalize the annotations.

% (150 pairs per dataset) to select the best judge model. This dataset includes randomly sampled pairs that were not matched by exact or relaxed matching, meaning they inherently represent challenging or ambiguous cases that simpler criteria could not resolve. Each pair is annotated as \textit{`match'} or \textit{`non-match'} by a human annotator. A second human verifies the labels, and disagreements are resolved through discussion to finalize the annotations.

\textbf{Judge LLM selection:} We evaluate both open-source (Llama3.1-70B) and closed-source (GPT-4o, GPT-4o-mini, GPT-3.5) models as potential judges, assessing their performance in \textit{zero-shot} and \textit{chain-of-thought} settings. GPT-4o, with a zero-shot prompt, achieves the highest agreement with human judgments, scoring 86.17. Therefore, we selected GPT-4o as the judge model for complex match evaluation. Note that the choice of judge is orthogonal to our proposed framework. The selected judge model can easily be swapped with newer or better alternatives without further modifications. Appendix \ref{appendix: judge-selection} provides additional details on judge selection and relevant prompts. 

\subsection{Judgment Alignment}
\label{judgment-alingment-analysis}
\begin{table}[h!]
\centering
\renewcommand*{\arraystretch}{0.9}
\small
\begin{tabular}{l|ccc|C{2.1cm}}
& \multicolumn{3}{c}{Deviation Rate (\%)}& Alignment (\%)\\

Datasets & EM & RM & CM & (1-$\sum$ deviation)\\
\midrule
DiscourseEE & 0.0 & 2.67 & 13.33 & 84.0 \\
PHEE &0.0 & 0.0 & 7.33 & 92.67\\
RAMS & 0.0 & 1.33 & 8.66 & 90.0\\
GENEVA & 0.0 & 2.0 & 8.0 & 90.0\\
DocEE & 0.0 & 3.33 & 16.0 & 80.67\\
WikiEvents &0.0 & 0.0 & 11.33 &  88.67\\
\midrule
&\multicolumn{3}{c}{Avg. Alignment (\%) } & 87.67\\
\bottomrule
\end{tabular}
\caption{Alignment and judgment deviation rate from humans at different matching levels on the evaluated EAE datasets.}
\label{judgement-error-rate}
\end{table}

We manually evaluated a subset of predictions from each matching level to determine the alignment with human judgments. In total, we analyzed 2,700 arguments (900 for each level) to quantify the frequency of disagreements with humans. To ensure unbiased judgments, we randomly selected 150 outputs from each level for each evaluated dataset. Table \ref{judgement-error-rate} presents the alignment and deviation rates. The EM consistently showed perfect alignment with human judgments, while RM exhibited minimal disagreement. However, CM demonstrated the highest deviation rates. 

Among the datasets, the PHEE dataset showed the highest alignment (92.67\%) with human judgments, while DocEE had the lowest (80.67\%). On average, the \textbf{REGen framework achieved 87.67\% alignment with human evaluators}. Our analysis reveals primary reasons for judgment disagreements are (1) \textit{Numerical nuances:} the model often failed to distinguish numerical differences. Such as for a role `\texttt{drug-dosage},' it incorrectly treated \texttt{`14 mg'} and \texttt{`6 mg'} as equivalent. (2) \textit{Temporal variations:} dates such as \texttt{`18 April'} versus \texttt{`20 April'} or days like \texttt{`Thursday'} versus \texttt{`Friday'} were incorrectly judged as similar. (3) \textit{Coreference handling:} datasets like RAMS and WikiEvents frequently used pronouns (e.g., `he', `they') in the ground truth, while models predicted specific names (e.g., `John'). This mismatch led to judgment errors, especially when documents contained multiple names, confusing the model. We identified a total of 111 disagreements. Of these, 15 cases were due to numerical nuances, 10 cases involved temporal variations, 57 cases were related to coreference handling, and 29 cases were due to other issues, such as the model incorrectly matching unrelated arguments. A detailed breakdown of these disagreement categories for each dataset is provided in Table \ref{table:error-breackdown}. Additional error analysis is provided in Appendix \ref{additional-error-analysis}.

\begin{table}[t!]
\centering
\renewcommand*{\arraystretch}{0.9}
\small
\begin{tabular}{l|C{1cm}C{1cm}C{1cm}C{1cm}}\\

& \multicolumn{4}{c}{Disagreement category}\\
\midrule
Datasets & Numerical & Temporal & Coreference & Other\\
\midrule
DiscourseEE & 7 & 1 & 15 & 1 \\
PHEE &1 & 1 & 4 & 5\\
RAMS & 0 & 0 & 10 & 5\\
GENEVA & 0 & 0 & 8 & 7\\
DocEE & 6 & 8 & 12 & 3\\
WikiEvents &1 & 0 & 8 & 8\\
\midrule
Total & 15 & 10 & 57 & 29\\
\bottomrule
\end{tabular}
\caption{Detailed breakdown of disagreement cases between human and judge model in evaluated EAE datasets.}
\label{table:error-breackdown}
\end{table}

The proposed JAM score accounts for these judgment errors. The score for each dataset is calculated based on alignment, providing a more reliable estimate of a model's true performance when using relaxed matching and LLM as judge models instead of human evaluators.

\section{Experiments}
\subsection{Datasets and Experimental Setup}
We used six standard EAE datasets from diverse domains to evaluate REGen. These datasets include: \textbf{RAMS} \cite{ebner-etal-2020-multi} (news), \textbf{GENEVA} \cite{parekh-etal-2023-geneva} (book, news, journal articles), \textbf{DocEE} \cite{tong-etal-2022-docee} (long news documents), \textbf{WikiEvents} \cite{li-etal-2021-document} (Wikipedia texts), \textbf{DiscourseEE} \cite{sharif-etal-2024-explicit} (online health discourse), and \textbf{PHEE} \cite{sun-etal-2022-phee} (pharmacovigilance texts). Prior works, such as \citet{huang-etal-2024-textee} and \citet{lu2024exactmatchsemanticallyreassessing} have evaluated LLMs using small test subsets sampled and merged from multiple datasets. Thus not reflecting actual performance of LLMs on these datasets. We conduct evaluations using the complete official test sets of the selected datasets to provide a more reliable assessment of LLMs' performance on these benchmarks. Detailed statistics for these test datasets are presented in Table \ref{data-statistics}. Appendix \ref{appendix: data-info} contains additional details on the data preparation steps.
%The same test set is used across all the models to ensure unbiased evaluation. 

\noindent
\textbf{Performance Metrics:} We report the precision, recall, and F1-score at each evaluation phase: exact match, relaxed match, complex match, and post-judgment alignment. Scores are computed following prior works \cite{peng-etal-2023-devil} and calculation details are discussed in Section \ref{mlm-eval-framework}.

\subsection{EAE Models}
\textbf{Baselines:} Following prior works \cite{sharif-etal-2024-explicit, lu-etal-2023-event}, we implement question-answering-based baselines. We use two models: BERT and FLAN-T5. Both models are fine-tuned on SQuAD \cite{rajpurkar-etal-2016-squad} data to extract \texttt{arguments} from \texttt{context} based on the \texttt{question}.  

\noindent
\textbf{LLM Based Models:} We perform comprehensive experiments using open-source and closed-source LLMs from different model families of various parameter sizes, including Phi-3.5 (3.8B), Gemma-1.1 (7B), Mixtral (8x7B), Llama-3.1 (70B), and GPT-4o. We evaluate all the models in two prompt settings: \textit{zero-shot} and \textit{chain-of-thought}. We employed question-guided prompting as previous works achieved SOTA performance using this approach \cite{lu-etal-2023-event,hsu-etal-2022-degree,du-cardie-2020-event}. Specifically, models are prompted with \texttt{(Instruction, Document, Question)} to generate  $\rightarrow$ \texttt{(Arguments)}, where each question is tailored to extract specific role\footnote{Role-specific questions for each dataset can be found here: \href{https://tinyurl.com/38en8e94}{https://tinyurl.com/38en8e94}}. Sample questions for the datasets are presented in Table \ref{role-specific-question-details}. 

Different LLMs require prompts and in-context samples tailored to each model and dataset. In practice, users select the optimal prompt using a trial-and-error approach \cite{ziems-etal-2024-large, 10.1145/3544548.3581388}. However, in our experiments, iterating over various prompts to find the optimal prompt for each model and dataset is impractical. Instead, we opted to use a consistent prompt across all models and datasets to (i) ensure a fair comparison among the models and (ii) eliminate the confounding factors related to prompt optimization.  Generic templates for zero-shot and chain-of-thought prompts for argument extraction are illustrated in Figures \ref{zs-EAE-prompt} and \ref{cot-EAE-prompt}, respectively. Additional descriptions of the models are provided in Appendix \ref{appendix: models}.

%We exclude few-shots and other prompt settings to reduce experimental complexity, noting that this can be explored in future work.
%\textcolor{blue}{[mention few shot as immediate future work as we could leverage insights from zero and CoT to design the few-shot experiments?]} {added in the limitations}
\section{Results}
\begin{table}[t!]
\centering
\renewcommand*{\arraystretch}{0.9}
\setlength{\tabcolsep}{4pt}
\small
\begin{tabular}{l|C{1cm}C{1.5cm}cC{1.2cm}}
Datasets& Avg. EM-F1 & Avg. REGen-F1 & $\Delta$ F1  &Gain (\%) \\
\toprule
DiscourseEE & 10.74 & 37.45 & +26.71 & +248.69 \\
PHEE       & 39.26 & 62.34 & +23.08 & +58.79 \\
RAMS       & 13.38 & 28.27 & +14.89 & +111.25 \\
GENEVA     & 13.62 & 46.12 & +32.49 & +238.52 \\
DocEE      & 17.33 & 41.65 & +24.32 & +140.36 \\
WikiEvents & 8.93  & 31.02 & +22.08 & +247.18 \\
\midrule
&\multicolumn{2}{c}{Avg. $\Delta$ F1}& +23.93 &\\
\bottomrule
\end{tabular}
\caption{Average F1-scores of LLMs in zero-shot and chain-of-thought settings, comparing Exact Match (EM) and REGen evaluation frameworks. Additional comparisons with Relaxed Match (RM) and Complex Match (CM) are reported in Table~\ref{all-comparison-F1}. } 
\label{average-improvement-rate}
\end{table}

\begin{table*}[t!]
\centering
\renewcommand*{\arraystretch}{1}
\small
\begin{tabular}{l|ccccC{1.5cm}C{1.3cm}c}
Datasets& \#Events & \#Roles & \#Docs  &  \#Arguments &  Doc-length (words) & Argument Density & Domain \\
\toprule
DiscourseEE & 3 & 34 & 98  & 997 & 121.21 & 10.17 &  Online health discourse\\
PHEE & 2 & 14 & 968  & 4952 & 20.12 & 5.11 & Pharmacovigilance \\
RAMS & 129 & 63 & 754  & 2023 & 133.70 & 2.68 & News \\
GENEVA & 115 & 196 & 899  & 3078 & 29.74 & 3.42 & General (book, news, journal)\\
DocEE & 57 & 266 & 500  & 3453 & 635.60 & 6.90 & News\\
WikiEvents & 33 & 44 & 19 & 473 & 653.87 & 24.89 &  Wikipedia\\
\bottomrule
\end{tabular}
\caption{Test set statistics of the six datasets used for evaluation show broad variability among these datasets. The columns \#Events, \#Roles, \#Docs, and \#Args represent the number of unique event types, unique role types, unique documents, and number of arguments, respectively. The average document length is measured in words, and argument density reflects the average number of arguments per document. } 
\label{data-statistics}
\end{table*}

\begin{table*}[h!]
\small
\centering
\renewcommand*{\arraystretch}{1}
\setlength{\tabcolsep}{5pt}
\begin{tabular}{l|cccc|cccc|cccc}
& \multicolumn{4}{c}{\textbf{DiscourseEE}} & \multicolumn{4}{c}{\textbf{PHEE}} & \multicolumn{4}{c}{\textbf{RAMS}} \\
\midrule
\textbf{Model}& EM & RM & CM & JAM & EM & RM & CM & JAM & EM & RM & CM & JAM \\
\midrule
\multicolumn{13}{c}{\textit{Baselines }}\\
\midrule
BERT & 5.88 & 8.66 & 33.56 & 30.18 & 27.78 & 34.98 & 52.61 & 51.33& 14.63 & 18.14 & 33.61 & 32.24 \\
Flan-T5 & 6.74 & 10.16 & 36.46 & 32.87 & 42.34 & 50.44 & 66.98 & 65.77 & 12.61 & 15.13 & 28.62 & 27.43\\
\midrule
\multicolumn{13}{c}{\textit{LLMs with Zero-Shot Prompt}}\\
\midrule

\includegraphics[width=0.30cm, height=0.30cm]{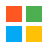} Phi-3.5  & 3.40 & 5.00 & 14.73 & 13.39 & 43.03 & 50.46 & 67.67 & 66.42 & 15.34 & 17.92 & 34.19 & 32.76 \\
\includegraphics[width=0.30cm, height=0.30cm]{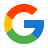} Gemma-1.1 & 11.87 & 15.86 & \cellcolor{green!50} 50.14 & \cellcolor{green!15} 45.48 & 45.00 & 54.34 & \cellcolor{green!15} 76.93 & \cellcolor{green!15} 75.28 & 14.87 & 17.50 & 32.43 & 31.11 \\
\includegraphics[width=0.30cm, height=0.30cm]{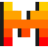} Mixtral & 13.10 & 17.74 & 48.59 & 44.38 & 36.58 & 42.55 & 59.19 & 57.98 & 12.97 & 15.46 & 29.93 & 28.65\\
\includegraphics[width=0.30cm, height=0.30cm]{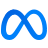} Llama-3.1  & 13.38 & 18.73 & 43.57 & 40.13 & 39.17 & 46.95 & 63.96 & 62.72 & 11.95 & 14.56 & 25.44 & 24.47  \\
\includegraphics[width=0.32cm, height=0.32cm]{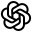} GPT-4o  & \cellcolor{green!50} 16.82 & \cellcolor{green!50} 23.08 & \cellcolor{green!15} 49.87 & \cellcolor{green!50} 46.16 & \cellcolor{green!50} 53.67 & \cellcolor{green!50} 61.92 & \cellcolor{green!50}78.96 &\cellcolor{green!50} 77.72 & \cellcolor{green!50}19.44 & \cellcolor{green!50} 23.15 & \cellcolor{green!50} 37.42 & \cellcolor{green!50} 36.15 \\

\midrule
\multicolumn{13}{c}{\textit{LLMs with Chain-of-thought Prompt}}\\
\midrule

\includegraphics[width=0.30cm, height=0.30cm]{icons/ms-icon.png} Phi-3.5  & 7.08 & 12.42 & 41.41 & 37.43 & 32.09 & 38.01 & 54.03 & 52.86 & \cellcolor{green!15} 15.53 & 18.65 & \cellcolor{green!15} 34.54 & \cellcolor{green!15} 33.13 \\
\includegraphics[width=0.30cm, height=0.30cm]{icons/google-icon.png} Gemma-1.1 & 9.35 & 13.06 & 43.27 & 39.16 & 34.14 & 42.28 & 61.46 & 60.06 & 10.71 & 13.57 & 26.28 & 25.15\\
\includegraphics[width=0.30cm, height=0.30cm]{icons/mistral-ai-icon.png} Mixtral & 4.99 & 7.00 & 26.39 & 23.76 & 29.46 & 37.28 & 50.75 & 49.77 & 6.85 & 8.28 & 17.00 & 16.23 \\
\includegraphics[width=0.30cm, height=0.30cm]{icons/meta-icon.png} Llama-3.1  & 12.65 & 17.31 & 44.75 & 40.98 & 31.29 & 39.93 & 52.51 & 51.59 & 10.66 & 12.76 & 23.72 & 22.75\\
\includegraphics[width=0.32cm, height=0.32cm]{icons/openai-icon.png} GPT-4o  & \cellcolor{green!15} 14.77 & \cellcolor{green!15} 20.86 & 47.33 & 43.66 & \cellcolor{green!15} 48.14 & \cellcolor{green!15} 55.66 & 70.01 & 68.96 & 15.50 & \cellcolor{green!15} 19.58 & 33.56 & 32.30 \\

\midrule
\midrule

& \multicolumn{4}{c}{\textbf{GENEVA}} & \multicolumn{4}{c}{\textbf{DocEE}} & \multicolumn{4}{c}{\textbf{WikiEvents}} \\
\midrule
\multicolumn{13}{c}{\textit{Baselines }}\\
\midrule
BERT & 15.24 & 26.58 & 53.09 & 50.74 & 18.66 & 25.74 & 47.81 & 44.05& 6.46 & 9.55 & 29.44 & 27.2 \\
Flan-T5 & \cellcolor{green!15} 18.34 & \cellcolor{green!15} 30.85 & \cellcolor{green!15} 57.76 & \cellcolor{green!15} 55.36 & 18.55 & 24.97 & 45.4 & 41.92 & 9.27 & 11.8 & 29.4 & 27.41 \\

\midrule
\multicolumn{13}{c}{\textit{LLMs with Zero-Shot Prompt}}\\
\midrule

\includegraphics[width=0.30cm, height=0.30cm]{icons/ms-icon.png} Phi-3.5  & 13.20 & 25.46 & 49.80 & 47.61 & 14.26 & 19.95 & 38.39 & 35.25 & 9.08 & 10.90 & 34.53 & 31.86 \\
\includegraphics[width=0.30cm, height=0.30cm]{icons/google-icon.png} Gemma-1.1 & 11.69 & 24.40 & 50.73 & 48.37 & 17.99 & 26.78 & 46.77 & 43.28 & 6.22 & 7.31 & 34.37 & 31.32 \\
\includegraphics[width=0.30cm, height=0.30cm]{icons/mistral-ai-icon.png} Mixtral & 13.31 & 24.86 & 48.44 & 46.32 & \cellcolor{green!50} 22.91 & \cellcolor{green!50} 32.55 & \cellcolor{green!50} 58.16 & \cellcolor{green!50} 53.74 & 9.89 & 12.36 & 38.24 &  35.31\\
\includegraphics[width=0.30cm, height=0.30cm]{icons/meta-icon.png} Llama-3.1  & 16.36 &  29.62 &  55.09 &  52.79 & 17.56 & 25.14 & 46.44 & 42.80 & \cellcolor{green!15} 12.81 & \cellcolor{green!15} 15.48 & 38.94 & 36.29 \\
\includegraphics[width=0.32cm, height=0.32cm]{icons/openai-icon.png} GPT-4o  & \cellcolor{green!50} 19.16 & \cellcolor{green!50} 33.35 & \cellcolor{green!50} 58.30 & \cellcolor{green!50} 56.02 & 21.91 & 31.65 & 56.40 & 52.14 & \cellcolor{green!50} 13.80 & \cellcolor{green!50} 17.00 & \cellcolor{green!15} 41.85 & \cellcolor{green!50}39.04\\

\midrule
\multicolumn{13}{c}{\textit{LLMs with Chain-of-thought Prompt}}\\
\midrule

\includegraphics[width=0.30cm, height=0.30cm]{icons/ms-icon.png} Phi-3.5  & 10.76 & 21.76 & 46.31 & 44.12 & 19.84 & 27.79 & 48.51 &  44.93 & 6.87 & 8.53 & 31.97 & 29.32 \\
\includegraphics[width=0.30cm, height=0.30cm]{icons/google-icon.png} Gemma-1.1 & 9.38 & 21.21 & 45.51 & 43.33 & 9.87 & 14.71 & 26.02 & 24.05 & 3.25 & 4.61 & 17.63 & 16.16 \\
\includegraphics[width=0.30cm, height=0.30cm]{icons/mistral-ai-icon.png} Mixtral & 16.37 & 26.77 & 47.07 & 45.24 & 6.90 & 9.63 & 19.28 & 17.65 & 4.49 & 6.07 & 19.58 & 18.06\\
\includegraphics[width=0.30cm, height=0.30cm]{icons/meta-icon.png} Llama-3.1  & 9.46 & 17.29 & 32.05 & 30.72 & 19.79 & 29.68 & 53.99 & 49.78 & 10.78 & 13.11 & 36.56 & 33.91 \\
\includegraphics[width=0.32cm, height=0.32cm]{icons/openai-icon.png} GPT-4o  &  16.54 & 28.50 & 48.48 & 46.65 & \cellcolor{green!15} 22.26 & \cellcolor{green!15} 31.92 & \cellcolor{green!15} 57.27 & \cellcolor{green!15} 52.90 & 12.15 & 15.10 & \cellcolor{green!50} 41.93 & \cellcolor{green!15} 38.90\\

\bottomrule
\end{tabular}

\caption{Evaluation results using the REGen framework for event argument extraction across the six datasets. The table reports F1-scores for models assessed at different evaluation levels: Exact Match (EM), Relaxed Match (RM), Complex Match (CM), and Judgment-Aligned Match (JAM). Due to space constraints, detailed precision, recall, F1-scores, and additional results are provided in Appendix Tables \ref{HM-improvement-rate}-\ref{WikiEvents-all-results}. The highest and the second-highest values in a column are highlighted using a dark shade and light shade, respectively.}
\label{EAE-all-results}
\end{table*}

\textbf{Significant improvement in F1-score across all datasets:} Table \ref{EAE-all-results} illustrates performance of various models using REGen framework. We observed a notable performance boost when models transitioned from the EM to the JAM score. For instance, the F1-score for the top-performing GPT-4o model increased from 16.82 with EM to 46.16 with JAM in the DiscourseEE dataset. Additionally, the average F1-scores of all the LLMs shown in Table \ref{average-improvement-rate} exhibit that all evaluated datasets achieved considerable performance gains, averaging 23.93 points. The increase in F1 score for the GENEVA dataset was 32.49, representing a 238.52\% improvement over the standard EM evaluation. Similar substantial gains were noted in other datasets, such as 26.71 for DiscourseEE and 24.32 for DocEE.

\begin{figure}[t!]
  \centering
  \includegraphics[width =1\linewidth]{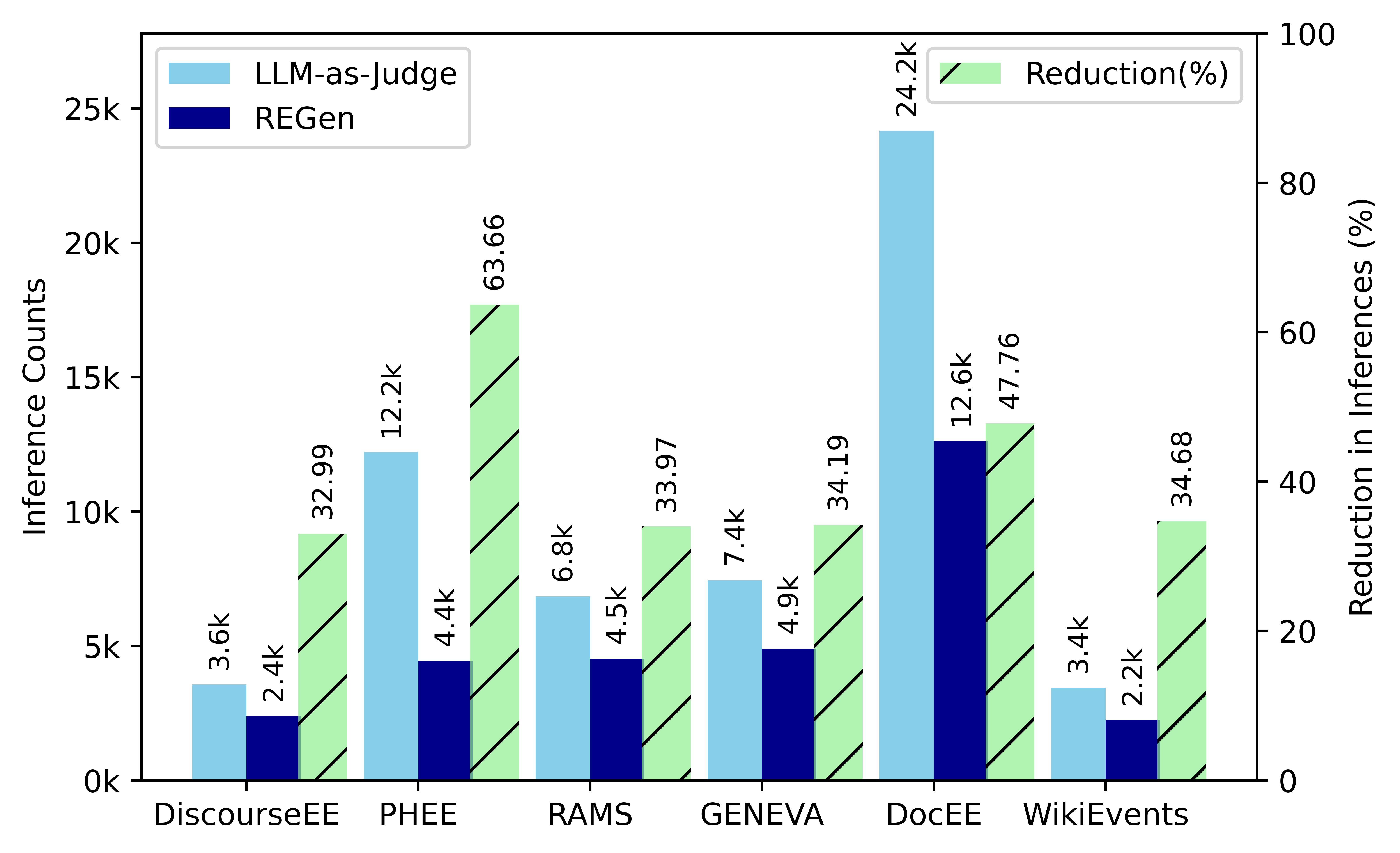}
 \caption{Comparison of required inference counts and reduction in inferences when using LLM-as-Judge versus the REGen framework for the GPT-4o prediction model. Additional statistics are presented in Table \ref{inference-reduction-full-stat}.}
 \label{inference-count-and-reduction-comparison}
\end{figure}

\textbf{On average, 41.20\% of inferences are reduced under the REGen framework:} Our results in Figure \ref{inference-count-and-reduction-comparison} and Table \ref{inference-reduction-full-stat} demonstrate that the REGen framework significantly lowers the number of inferences needed for evaluation compared to solely using the LLMs-as-judge approach \cite{lu2024exactmatchsemanticallyreassessing}. For example, in the PHEE dataset, the inference count drops dramatically from 12,206 to 4,436, resulting in a reduction of 63.6\%. Similarly, the DocEE dataset sees a decrease from 24,166 to 12,624, corresponding to a 47.7\% reduction. These results highlight the efficiency of the REGen framework in streamlining the inference process. It enables effective evaluation by significantly decreasing the computational burden. Moreover, the systematic reduction in judgment errors through the REGen framework lessens the need for human validation without compromising reliability.

\textbf{REGen framework is more reliable (87.67\% alignment)}: REGen shows no/minimal errors in the performance under exact match and relaxed match scoring. While there is some overestimation due to misjudgments in the complex match step, our extensive validation indicates an 87.67\% alignment with human judgments (see Table \ref{judgement-error-rate}). The JAM score incorporates this human alignment, ensuring the overall reliability of the framework. Additionally, the reported scores are more explainable, as they include a clear breakdown of performance gains at each matching level (EM, RM, CM, and JAM).

\textbf{Recall is on average higher than precision in all settings:} Our fine-grained analysis (see Tables \ref{DiscourseEE-all-results} to \ref{WikiEvents-all-results}) reveals LLMs achieve higher recall than precision. Such as the GPT-4o model in the DocEE dataset achieved a JAM recall of 68.41 compared to a precision of only 42.12. This indicates while the models are effective in identifying ground-truth arguments, they tend to over-predict, impacting the overall F1-score. In this work, we used a single prompt for all the models and datasets, which might have contributed to this overprediction. Future research should focus on pushing the performance through dataset- and model-specific prompting to enhance precision without sacrificing recall.

\section{Related Work}
\textbf{Generative Event Argument Extraction:} 
Early studies on event argument extraction (EAE) treated it as an extractive or token-level classification task \cite{doddington-etal-2004-automatic, du-cardie-2020-event}. These efforts primarily focused on identifying argument spans directly found in the text \cite{sun-etal-2022-phee}. Recently, EAE has been formulated as a generative task where pre-trained language models are guided with natural language to fill templates or generate arguments \cite{hsu-etal-2022-degree}. \citet{sharif-etal-2024-explicit} argue that this generative formulation better suits real-world applications as it can capture implicit and scattered arguments better. With the emergence of LLMs,  generative model-based argument extraction gained more traction \cite{sun-etal-2024-leveraging, he-etal-2024-demonstration, gatto-etal-2025-document}. So, we focus on generative extraction covering diverse models and datasets. 

%where the predicted argument must exactly match the ground-truth span
\noindent
\textbf{Evaluations for Generative EAE:} Existing works for generative EAE primarily rely on \textit{exact matching} for evaluation \cite{huang-etal-2024-textee}. This strict approach unfairly penalizes models, even when the generated output is correct \cite{fane-etal-2025-bemeae}. To address this, \citet{han2024empiricalstudyinformationextraction} adopts a relaxed matching approach, considering arguments similar if their embedding-based similarly exceeds a threshold of 0.5. Similarly, \citet{sharif-etal-2024-explicit} used a threshold of 0.75. However, this approach has limitations. It fails to capture semantically similar arguments with different lexical or syntactic forms and wrongly classifies arguments with high token overlap as similar. Thus, performance reported solely on relaxed matching is unreliable. More recently, 
\citet{lu2024exactmatchsemanticallyreassessing} employed LLMs to determine argument similarity. Nonetheless, this approach incurs significant computational overhead and demands extensive human validation. Our REGen framework combines the strengths of exact, relaxed, and LLM-based matching. It systematically reduces misjudgments, computational costs, and the need for human validation. 
% Finally, the judgment alignment component ensures reliable performance evaluation. 
\section{Conclusion}
This paper presents REGen, a novel evaluation framework for EAE. Our extensive experiments and human validation demonstrate its effectiveness, with a 23.93-point gain in average F1 score across six EAE datasets, and reliability, with 87.67\% alignment with human judgments. We highlight the limitations of current evaluation approaches and illustrate how REGen addresses these issues. Furthermore, our analysis reveals that previous studies have underestimated the true performance of LLMs. We believe that REGen fills a critical gap in EAE research and motivates future work to explore the generative model's capability in solving other information extraction tasks, e.g., relation extraction, entity extraction, and beyond.

\section{Limitations}
One limitation of this work is that we did not conduct statistical significance testing on the reported results.  We chose not to conduct statistical testing for two reasons. First, our goal is not to conclude which model is best but to highlight performance gaps and show how existing evaluation approaches underestimate model performance. The results clearly demonstrate a significant performance gap with exact match evaluation, which is not diminished by the lack of statistical testing. Second, performing statistical tests across all datasets and models with multiple runs is time-consuming and prohibitively expensive. For example, averaging over 3 runs would require an additional 320k inferences.

Another limitation is that we did not optimize prompts for each model. Performance could be improved with dataset- and model-specific prompting. However, we chose to focus on benchmarking a wide range of datasets using model-agnostic prompting. Conducting a thorough, prompt engineering for every model and dataset in a single study is not feasible. Our results show significant performance gains and future work can explore dataset- and model-specific prompts to further enhance performance. Additionally, future work can explore few-shot experiments and find optimal prompting strategies for different datasets, such as self-consistency \cite{wang2023selfconsistencyimproveschainthought} or plan-and-solve \cite{wang-etal-2023-plan}. We exclude few-shot experiments as they require selecting demonstration examples through trial and error, finding the optimal order of demonstration, and running multiple iterations, which significantly increases the experimental cost and complexity.

\section*{Ethical Considerations}

\textbf{Intended Use:} We released our judge and alignment datasets to facilitate future research on generative argument extraction evaluation.

\noindent
\textbf{Annotation:} Judge and alignment data annotation were conducted by trained NLP researchers. All annotators were compensated as per the standard paying rate of the author's institution. Key characteristics of our annotators include: (a) graduate students, (b) 3-6 years of research experience, and (c) a mix of native and non-native English speakers. We provided annotators detailed annotation guidelines, including argument extraction, semantical similarity, and the type of information we wanted to compare to mitigate potential biases. 

\noindent
\textbf{Reproducibility} Details on models and dataset processing are provided in Appendices \ref{appendix: data-info} and \ref{appendix: models}. The evaluation framework, code, and processed datasets are available at \href{https://github.com/Omar-Sharif/REGen}{https://github.com/Omar-Sharif/REGen}..

\section*{Acknowledgments}
This work was supported in part by National Institutes of Health (NIH) grant
1R21DA059665-01A1. We also thank the reviewers for their valuable feedback.

\bibliography{acl_latex}

\begin{table*}[t!]
\centering
\small
\renewcommand*{\arraystretch}{1.1}
\begin{tabular}{L{2.3cm}L{3cm}L{3.7cm}C{1.4cm}C{1.4cm}C{1.5cm}}

\toprule
\textbf{Role} & \textbf{Ground-truth} & \textbf{Prediction} [variations] & \textbf{Exact Match} & \textbf{Relaxed Match} & \textbf{Complex Match} \\
\midrule 
 % & patch & patch & Match | \includegraphics[width=0.35cm, height=0.35cm]{icons/green-tick.png} &  Match | \includegraphics[width=0.30cm, height=0.30cm]{icons/green-tick.png} & Match | \includegraphics[width=0.30cm, height=0.30cm]{icons/green-tick.png} \\
 % & the patch & Not match | \includegraphics[width=0.30cm, height=0.30cm]{icons/red-cross.png} & Match | \includegraphics[width=0.35cm, height=0.35cm]{icons/green-tick.png} &  Match | \includegraphics[width=0.35cm, height=0.35cm]{icons/green-tick.png}\\
 % & the transdermal patch & Not match | \includegraphics[width=0.30cm, height=0.30cm]{icons/red-cross.png} & Not match | \includegraphics[width=0.30cm, height=0.30cm]{icons/red-cross.png} & Match | \includegraphics[width=0.35cm, height=0.35cm]{icons/green-tick.png} \\
 % \midrule 
 \texttt{Date} & 20 April 2024 & 20 April 2024 & M | \includegraphics[width=0.35cm, height=0.35cm]{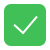} &  M | \includegraphics[width=0.30cm, height=0.30cm]{icons/green-tick.png} & M | \includegraphics[width=0.30cm, height=0.30cm]{icons/green-tick.png} \\

 & & 18 April 2024 & NM | \includegraphics[width=0.30cm, height=0.30cm]{icons/green-tick.png} & M | \includegraphics[width=0.30cm, height=0.30cm]{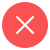} & NM | \includegraphics[width=0.30cm, height=0.30cm]{icons/green-tick.png}\\ 

 & & 17 December 2022 & NM | \includegraphics[width=0.30cm, height=0.30cm]{icons/green-tick.png} & NM | \includegraphics[width=0.30cm, height=0.30cm]{icons/green-tick.png}  & NM | \includegraphics[width=0.30cm, height=0.30cm]{icons/green-tick.png} \\
\midrule

\texttt{Cancer status} & Cancer has spread & Patient cancer has spread & NM | \includegraphics[width=0.30cm, height=0.30cm]{icons/red-cross.png} & M | \includegraphics[width=0.30cm, height=0.30cm]{icons/green-tick.png} & M | \includegraphics[width=0.30cm, height=0.30cm]{icons/green-tick.png} \\
& & Metastasized & NM | \includegraphics[width=0.30cm, height=0.30cm]{icons/red-cross.png} & NM | \includegraphics[width=0.30cm, height=0.30cm]{icons/red-cross.png} & M | \includegraphics[width=0.30cm, height=0.30cm]{icons/green-tick.png} \\
& & Cancer spread from lung to liver & NM | \includegraphics[width=0.30cm, height=0.30cm]{icons/red-cross.png} & NM | \includegraphics[width=0.30cm, height=0.30cm]{icons/red-cross.png} & M | \includegraphics[width=0.30cm, height=0.30cm]{icons/green-tick.png} \\
\midrule

\texttt{Occasion} & July 4th & US Independence Day & NM | \includegraphics[width=0.30cm, height=0.30cm]{icons/red-cross.png} & NM | \includegraphics[width=0.30cm, height=0.30cm]{icons/red-cross.png} & M | \includegraphics[width=0.30cm, height=0.30cm]{icons/green-tick.png}\\
 &  & Fourth of July & NM | \includegraphics[width=0.30cm, height=0.30cm]{icons/red-cross.png} & M | \includegraphics[width=0.30cm, height=0.30cm]{icons/green-tick.png} & M | \includegraphics[width=0.30cm, height=0.30cm]{icons/green-tick.png}\\
 \midrule

 \texttt{Participant} & John Kerry & Senator John Kerry& NM | \includegraphics[width=0.30cm, height=0.30cm]{icons/red-cross.png} & M | \includegraphics[width=0.30cm, height=0.30cm]{icons/green-tick.png} & M | \includegraphics[width=0.30cm, height=0.30cm]{icons/green-tick.png}\\
& & US secretary of state in 2014 & NM | \includegraphics[width=0.30cm, height=0.30cm]{icons/red-cross.png} & NM | \includegraphics[width=0.30cm, height=0.30cm]{icons/red-cross.png} & M | \includegraphics[width=0.30cm, height=0.30cm]{icons/green-tick.png}\\
\midrule

\texttt{Drug dosage} & 6 mg & 6 mg & M | \includegraphics[width=0.35cm, height=0.35cm]{icons/green-tick.png} &  M | \includegraphics[width=0.30cm, height=0.30cm]{icons/green-tick.png} & M | \includegraphics[width=0.30cm, height=0.30cm]{icons/green-tick.png} \\
&& 14 mg & NM \includegraphics[width=0.35cm, height=0.35cm]{icons/green-tick.png} & M | \includegraphics[width=0.30cm, height=0.30cm]{icons/red-cross.png} & M | \includegraphics[width=0.30cm, height=0.30cm]{icons/red-cross.png}\\
&& 6 milligram & NM | \includegraphics[width=0.30cm, height=0.30cm]{icons/red-cross.png} & NM | \includegraphics[width=0.30cm, height=0.30cm]{icons/red-cross.png} & M | \includegraphics[width=0.30cm, height=0.30cm]{icons/green-tick.png}\\ 
&& take 6 mg drug & NM | \includegraphics[width=0.30cm, height=0.30cm]{icons/red-cross.png} & NM | \includegraphics[width=0.30cm, height=0.30cm]{icons/red-cross.png} & M | \includegraphics[width=0.30cm, height=0.30cm]{icons/green-tick.png}\\ 
\midrule

% \texttt{Duration} & 1 month & 2 month & NM | \includegraphics[width=0.35cm, height=0.35cm]{icons/green-tick.png} &  M | \includegraphics[width=0.30cm, height=0.30cm]{icons/red-cross.png} & M | \includegraphics[width=0.30cm, height=0.30cm]{icons/red-cross.png} \\
\texttt{Duration} & 1 month & from July 30 to August 30 & NM | \includegraphics[width=0.30cm, height=0.30cm]{icons/red-cross.png} & NM | \includegraphics[width=0.30cm, height=0.30cm]{icons/red-cross.png} & M | \includegraphics[width=0.35cm, height=0.35cm]{icons/green-tick.png}\\
& & 30 days & NM | \includegraphics[width=0.30cm, height=0.30cm]{icons/red-cross.png} & NM | \includegraphics[width=0.30cm, height=0.30cm]{icons/red-cross.png} & M | \includegraphics[width=0.35cm, height=0.35cm]{icons/green-tick.png}\\
& & One month & NM | \includegraphics[width=0.30cm, height=0.30cm]{icons/red-cross.png} & M | \includegraphics[width=0.35cm, height=0.35cm]{icons/green-tick.png} & M | \includegraphics[width=0.35cm, height=0.35cm]{icons/green-tick.png}\\
\midrule

\texttt{Medical condition} & Chronic kidney disease & Long term kidney disease & NM | \includegraphics[width=0.30cm, height=0.30cm]{icons/red-cross.png} &  NM | \includegraphics[width=0.30cm, height=0.30cm]{icons/red-cross.png} & M | \includegraphics[width=0.35cm, height=0.35cm]{icons/green-tick.png} \\
&& long term heart disease & NM | \includegraphics[width=0.30cm, height=0.30cm]{icons/green-tick.png} &  NM | \includegraphics[width=0.30cm, height=0.30cm]{icons/green-tick.png} & NM | \includegraphics[width=0.35cm, height=0.35cm]{icons/green-tick.png} \\
&& CKD & NM | \includegraphics[width=0.30cm, height=0.30cm]{icons/red-cross.png} &  NM | \includegraphics[width=0.30cm, height=0.30cm]{icons/red-cross.png} & NM | \includegraphics[width=0.35cm, height=0.35cm]{icons/red-cross.png} \\
\midrule

\texttt{Causalities and losses} & 7 dead in central provinces flooding & 18 died in central provinces due to flood & NM | \includegraphics[width=0.30cm, height=0.30cm]{icons/green-tick.png} &  M | \includegraphics[width=0.30cm, height=0.30cm]{icons/red-cross.png} & NM | \includegraphics[width=0.35cm, height=0.35cm]{icons/green-tick.png} \\

&& Flooding in the central provinces killed seven people & NM | \includegraphics[width=0.30cm, height=0.30cm]{icons/red-cross.png} &  M | \includegraphics[width=0.30cm, height=0.30cm]{icons/green-tick.png} & M | \includegraphics[width=0.35cm, height=0.35cm]{icons/green-tick.png} \\
\midrule

\texttt{Place} & Chelsea, New York & Chelsea neighborhood in NYC &  NM | \includegraphics[width=0.30cm, height=0.30cm]{icons/red-cross.png} &  NM | \includegraphics[width=0.30cm, height=0.30cm]{icons/red-cross.png} & M | \includegraphics[width=0.35cm, height=0.35cm]{icons/green-tick.png}\\
&& Chelsea, NYC & NM | \includegraphics[width=0.30cm, height=0.30cm]{icons/red-cross.png} & M | \includegraphics[width=0.30cm, height=0.30cm]{icons/green-tick.png} & M | \includegraphics[width=0.30cm, height=0.30cm]{icons/green-tick.png}\\

&& Manhattan & NM | \includegraphics[width=0.30cm, height=0.30cm]{icons/green-tick.png} & NM | \includegraphics[width=0.30cm, height=0.30cm]{icons/green-tick.png} & M | \includegraphics[width=0.30cm, height=0.30cm]{icons/red-cross.png}\\

\bottomrule
\end{tabular}
 
\captionof{table}{\label{error-visualization} Illustration of how predicted arguments may differ from ground-truth arguments for a specific role and how they are evaluated under different approaches—exact match, relaxed match, and complex match. Each cell shows whether the argument pair is classified as a match (M) or not a match (NM) under the respective method. Green checkmarks ($\checkmark$) indicate correct evaluations, while red crosses ($X$) indicate errors. Our analysis shows that relying on a single evaluation method results in inaccurate assessments. REGen addresses this by systematically combining the strengths of each approach, enhancing evaluation accuracy while reducing computational cost.}
\end{table*}

\begin{table*}[h!]
\centering
\small
\renewcommand*{\arraystretch}{1}
\begin{tabular}{L{1.8cm}L{2.5cm}L{2.5cm}L{2.5cm}L{4.5cm}}

\midrule
 \textbf{Role} & \textbf{Query} & \textbf{Ground-truth} & \textbf{Prediction} & \textbf{Observations}\\
\midrule
\multicolumn{5}{c}{\textit{Example 1}}\\

\toprule
\multicolumn{5}{L{15.5cm}}{I'm \colorbox{cyan!20}{46 years} old and recently diagnosed with prostate cancer, which my doctor described as moderately aggressive. The doctor recommended \colorbox{green!20}{robotic surgery} using the \colorbox{green!20}{da Vinci system} due to its precision and faster recovery, but I'm hesitant because of possible long-term side effects. [...]. My \colorbox{brown!20}{cancer has already spread} to nearby lymph nodes, which makes me feel like my options are shrinking. [..]}\\
\midrule

\texttt{Treatment options} &  What treatment options patient have? & \cellcolor{green!20} da Vinci robotic surgery & robotic assisted surgery & Scattered arguments; Arguments refer to the same treatment options.\\
\midrule

\texttt{Age} &  What is the age of the patient? &  \cellcolor{cyan!20} 46 years & 46 & Here, 46 refers to patient age. Exact and relaxed match approach fails to correlate.\\
\midrule
\texttt{Cancer status} &  What is patient cancer status? &  \cellcolor{brown!20} Cancer has spread & Metastasized & Semantically similar arguments based on context despite lexical and syntactic differences.\\
\midrule

\midrule
\multicolumn{5}{c}{\textit{Example 2}}\\
\midrule
\multicolumn{5}{L{15.5cm}}{On March 15, a group of militants launched a \colorbox{orange!20}{coordinated assault} on a military outpost in northern Mali. The surprise attack, believed to be conducted by a terrorist group linked to \colorbox{magenta!20}{al-Qaeda (AQ)}, involved \colorbox{teal!20}{several trucks and heavy gunfire}. The attack resulted in the deaths of \colorbox{yellow!20}{17 military personnel}. Government officials reported that the AQ seized control of the outpost before reinforcements could arrive.}\\
\midrule

\texttt{Attack type} & What type of attack occurred? & \cellcolor{orange!20} coordinated assault & surprise attack & Example of subjective annotation. Both arguments can be accurate based on the role.\\
\midrule
\texttt{Attacker} & Who carried out the attack? & \cellcolor{magenta!20} al-Qaeda & AQ & Coreference: referring to same entity\\

\midrule
\texttt{Attack Weapon} & What weapons were used in the attack? & \cellcolor{teal!20}several trucks and guns &  trucks, guns & Difference in span boundary\\

\midrule
\texttt{Causalities} & How many were killed? & \cellcolor{yellow!20} 17 military personnel & 17 militants & Lexical variation\\

\bottomrule
\end{tabular}
 
\captionof{table}{\label{error analysis} Examples illustrating the effectiveness of REGen framework. In these cases, existing evaluation approaches (Exact match, Relaxed match, Head noun phrase match) fail to capture the semantic similarity between ground-truth and predicted arguments. For each example, representative argument roles are shown. Highlighted colors indicate the source segment of the ground-truth annotation.}
\end{table*}
\newpage

\appendix
\section*{Appendix}
\section{Error Analysis}
\label{additional-error-analysis}
Table \ref{error-visualization} presents examples of generative models' prediction variations from the ground-truth arguments for a specific role. It also demonstrates how these argument pairs are evaluated under exact-match, relaxed-match, and complex-match evaluation schemes. For instance, consider the role \texttt{Cancer status} for which the ground-truth and predicted arguments are \texttt{`cancer has spread'} and \texttt{`metastasized'}. The SBERT similarity score of these two arguments is 0.1597. But based on the context (document, event, role), these two arguments should be considered a match, and both exact and relaxed-match fail to recognize this. This causes severe underestimation of model performance and unfair evaluation. As shown in the table, complex matching performs well in such cases. However, the complex match can sometimes fail to distinguish differences. Such as for the role \texttt{Place}, it incorrectly treated \texttt{`Chelsea, New York'} and \texttt{`Manhattan'} as similar. The complex match model might consider them as similar because Manhattan is part of New York, and Chelsea is part of Manhattan. We further discuss these errors in Section \ref{judgment-alingment-analysis}. The REGen framework hierarchically combines complex matching with exact, relaxed matching, and judgment alignment, helping capture true model performance. We illustrate the need for a context-grounded evaluation approach and the effectiveness of the REGen framework through examples in Table \ref{error analysis}.

\begin{figure*}[t!]
  \centering
  \includegraphics[width =1\linewidth]{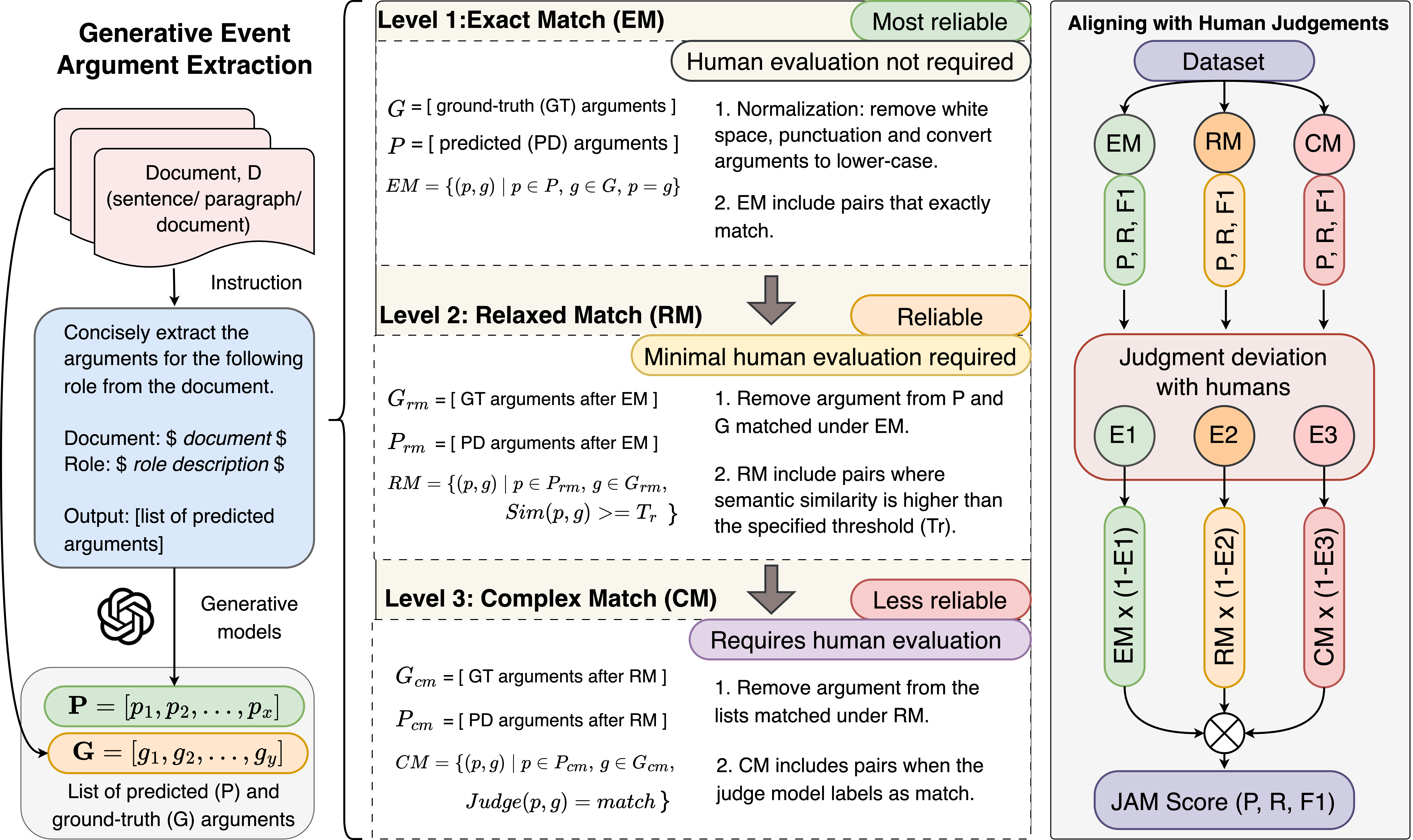}
 \caption{Illustration of REGen evaluation framework. \textbf{\textit{Left:}} Example showing getting role-specific arguments from documents using generative models. \textbf{\textit{Middle:}} Evaluation process at different levels: Exact Match, Relaxed Match, and Complex Match. \textbf{\textit{Right:}} Judgment Aligned Match (JAM) score calculation process on a specific dataset, where E1, E2, and E3 represent the deviation rates from human judgments at different matching levels. Typically, E1 equals zero for a dataset, as an exact match indicates a perfect agreement with humans. \textit{P, R,} and \textit{F1} denote Precision, Recall, and F1-score, respectively. JAM score ensures that reported scores are reliable and not inflated due to misjudgments.}
 \label{eval-framework}
\end{figure*}

\section{Alignment Calculation Details}
\label{appendix:alignment-calculation}
In the judgment alignment calculation (section \ref{subsection:judgement-alignment}) we count disagreement ($N_d$) only when the judge model marks a prediction as a match, but the human annotator marks it as a non-match. This choice helps mitigate issues of over-penalization and overestimation.

\begin{itemize}
    \item \textbf{Over-penalization issue:} The precision and recall for argument evaluation are calculated differently than in standard classification tasks. Specifically, there are no true negative cases because the total number of negative arguments is unknown \cite{sharif-etal-2024-explicit,sun-etal-2022-phee}. Instead, precision measures how many predicted arguments are correct, and recall measures how many ground-truth arguments were correctly identified by the model (Sections \ref{subsection:exact-match}–\ref{subsection:complex-match}). These two numbers can vary because multiple predicted arguments can be matched with a single ground-truth argument and vice versa. Because the score is computed only based on matches, any argument pairs predicted as non-matches do not contribute to the final score. Therefore, cases where the judge model predicts a non-match but a human annotator marks it as a match do not affect the evaluation score. Including such cases in the disagreement count will increase the deviation rate and would unfairly penalize the model’s performance during JAM score calculation. To prevent over-penalization, we exclude these cases from our disagreement rate.

    \item \textbf{Ensuring no overestimation:} To obtain the most accurate evaluation score, all evaluations were made by the relaxed match model, and the complex match judge needs to be evaluated by humans. It is not feasible to conduct all the reevaluations manually. Deviation rate helps us obtain a reliable estimate of model performance. There is a trade-off – if relaxed and complex match criteria are too strict, we risk underestimation; if they are too lenient, we risk overestimation. To address this, we prioritize precision: when our framework says there is a match, we want to be confident it is indeed a true match. By factoring the score with the deviation rate, the final score we report is intentionally conservative — it serves as a lower bound on the model’s true performance, avoiding inflated scores due to unverified matches.
\end{itemize}
\section{Dataset Details}

\label{appendix: data-info}
We transform all datasets into a unified format as explained in Section \ref{dataset-formatting}. Each document includes a set of predefined roles based on the event, with each role having a list of ground-truth argument strings. In this work, we use a trigger-free approach for argument extraction \cite{tong-etal-2022-docee}. We adopt this formulation because many datasets lack trigger annotations or include implicit or scattered arguments that can not be tied to trigger phrases \cite{sharif-etal-2024-explicit}. We use the official test split of all the datasets. Table \ref{data-statistics} exhibits the detailed statistics of the datasets. We will release our processed datasets and associated scripts upon acceptance of the paper. Detailed descriptions of each dataset are provided in the following.

\begin{itemize}
    \item  \textbf{DiscourseEE} \cite{sharif-etal-2024-explicit} dataset is annotated from online health discussions and includes explicit, implicit, and scattered arguments. This dataset is hierarchical, with each role further classified into four types: core, type-specific, subject-specific, and effect-specific arguments. We sourced the test set from the official repository 
\href{https://github.com/omar-sharif03/DiscourseEE/tree/main/Data}{https://github.com/omar-sharif03/DiscourseEE}. It features 34 unique roles across 3 event types, with all arguments annotated as strings. In this work, we do not use the argument types or hierarchical structure, as they are not essential. We process the dataset using the author's provided code.

\item  \textbf{PHEE} \cite{sun-etal-2022-phee} is an event extraction dataset sourced from the pharmacovigilance domain. It contains 14 unique roles across 2 event types. We obtain the dataset from \href{https://github.com/ZhaoyueSun/PHEE}{https://github.com/ ZhaoyueSun/PHEE}. The dataset includes annotations for both trigger and argument spans. Following our formulation, we discard the trigger and only take the argument strings. We combine multiple arguments under the same role into a single argument list, separating them with semicolons.

\item \textbf{RAMS} \cite{ebner-etal-2020-multi} is an event extraction dataset from the news domain. We downloaded the dataset from \href{https://nlp.jhu.edu/rams/}{https://nlp.jhu.edu/rams/} and processed it leveraging the script provided by TextEE \cite{huang-etal-2024-textee}. We ignored the trigger annotation and used the argument string to map the dataset into our formulation. The test set contains 129 unique events and 63 roles.

\item \textbf{GENEVA} \cite{parekh-etal-2023-geneva} is a general-domain event extraction dataset developed using FrameNet. This dataset includes samples from books, articles, journals, and Wikipedia. We used the provided test set from \href{https://github.com/PlusLabNLP/GENEVA/tree/main/data}{https://github.com/PlusLabNLP/GENEVA}. The test set includes 115 events and 196 unique roles. We applied the preprocessing script from TextEE \cite{huang-etal-2024-textee} to convert the dataset to our format.

\item \textbf{DocEE} \cite{tong-etal-2022-docee} is a trigger-free document-level event extraction dataset with very long documents. We obtained the test set from the official GitHub repository \href{https://github.com/tongmeihan1995/DocEE}{https://github.com/tongmeihan1995/DocEE}. The official test set contains 2,771 documents. Due to high inference time, we selected 500 random samples to reduce complexity. Our test set includes 57 unique events and 266 unique roles.

\item \textbf{WikiEvents}: \cite{li-etal-2021-document} is a document-level event extraction dataset based on Wikipedia texts. We sourced the dataset from \href{https://github.com/raspberryice/gen-arg}{https://github.com/raspberryice/gen-arg} and processed it using the TextEE \cite{huang-etal-2024-textee} preprocessing script. We retained only the argument annotations and discarded the rest. The test set includes 33 event types and 44 roles. WikiEvents is highly argument-dense compared to other datasets, with a density of 24.89. It also has longer documents, averaging 654 words per document.

\end{itemize}

\section{Model Details}
\label{appendix: models}
\textbf{Baselines:} As baselines, we used transformer-based BERT (110M parameters) and instruction-fine-tuned FLAN-T5 (250M parameters) models. Both models were implemented using the HuggingFace pipeline and fine-tuned on the SQuAD \cite{rajpurkar-etal-2016-squad} dataset. BERT was fine-tuned with a learning rate of $2 \times 10^{-5}$, a batch size of 8, and trained for 3 epochs. FLAN-T5 was fine-tuned for 4 epochs with a batch size of 16 and the same learning rate. During prediction, we provide a role-specific question and the associated document as context. The input is formatted as \texttt{[CLS] Question [SEP] Document [SEP]}. The output span is then decoded as the argument for the specific role. Arguments for each role are extracted independently.

\paragraph{LLMs:} To investigate the feasibility of our proposed evaluation framework, we experimented with various LLMs used in previous studies on event argument extraction \cite{sharif-etal-2024-explicit,lu2024exactmatchsemanticallyreassessing}. We evaluated open-source models ranging from 4B to 70B parameters and the closed-source GPT-4o model. This diverse selection allowed us to assess performance across different scales. We used five LLMs for the experimentation.
\begin{itemize}
    \item \textbf{Phi-3.5} : 
    We used the Phi-3.5-mini, a 3.8 billion parameter model trained on 3.3 trillion tokens \cite{phi-3.5}. It achieves comparable performance to Mixtral  8x7B and GPT-3.5 models on academic benchmarks despite being a very small model. 
    
    \item \textbf{Gemma-1.1} model trained on 6T tokens with novel RLHF method, based on the architecture and training recipe of Gemini models \cite{gemmateam2024gemma}. It performs better than similar open-source models in 11 out of 18 text tasks. Gemma is available in two versions, with 2 billion and 7 billion parameters. We use the 7 billion parameter version for our experiments.
    
    \item \textbf{Mixtral (8x7B)} is a sparse mixture of expert language designed with an architecture similar to Mistral 7B \cite{jiang2024mixtral}. It has a total of 47 billion parameters, with only 13 billion being active at a time. These architectural changes allow Mixtral to outperform models with more parameters (e.g., Llama-2, GPT-3.5) across several benchmarks. 
    
    \item \textbf{Llama-3.1} is a state-of-the-art open-source language model pretrained and instruction-fined with 8B, 70B, and 405B parameters. It builds upon the Llama-3 model \cite{llama-3.1}, incorporating grouped query attention (GQA) and RLHF. We use the 70B version of the model.

    \item \textbf{GPT-4o} \cite{openai2024gpt4} is one of the best-performing models that can reason across audio, vision, and text. It achieved state-of-the-art performance across most benchmarks\footnote{https://lmarena.ai/}. 
\end{itemize}

We utilized the instruction-tuned versions of all the models. The HuggingFace inference strings for the open-source LLMs are Phi-3.5 \texttt{(microsoft/ Phi-3.5-mini-instruct)}, Gemma-1.1 \texttt{(google/ gemma-1.1-7b-it)}, Mixtral \texttt{(mistralai/ Mixtral-8x7B-Instruct- v0.1)}, and Llama-3.1 \texttt{(meta-llama/ Llama-3.1-70B-Instruct)}. We assess the performance of the GPT-4o model through API calls, using version \texttt{(gpt-4o-2024-11-20)}.
\section{Judge Selection Process}
\label{appendix: judge-selection}
\begin{figure}[t!]
  \centering
  \includegraphics[width =1\linewidth]{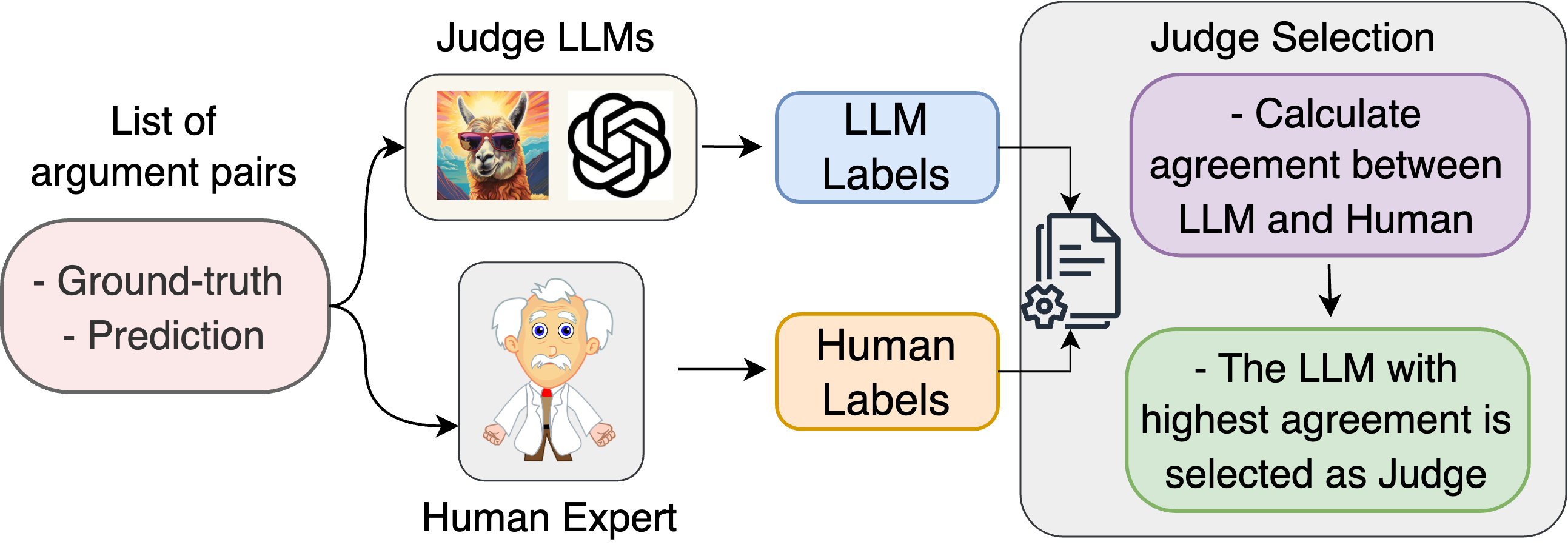}
 \caption{Schematic diagram of the judge selection process. }
 \label{judge-selection-process}
\end{figure}

There is a growing trend to leverage LLM as a judge to reduce the high cost of human evaluation \cite{NEURIPS2023_91f18a12, GPTEval, gu2025surveyllmasajudge}. Following this approach, we employed LLMs to mimic human evaluation and automatically determine whether the ground truth and predicted arguments match. Figure \ref{judge-selection-process} shows the schematic diagram of the judge selection process. 

\begin{figure}[h!]
  \centering
  \includegraphics[width =1\linewidth]{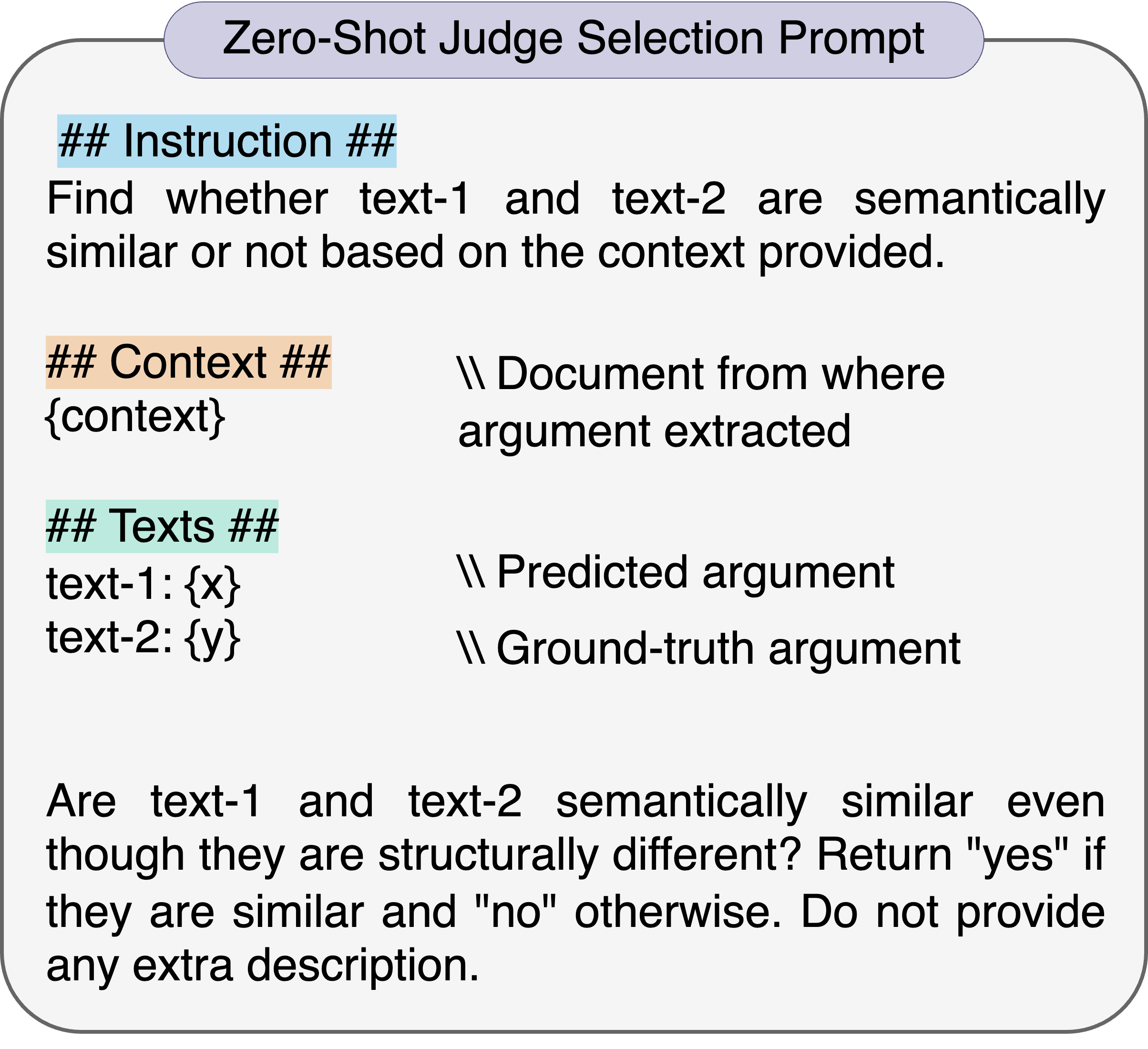}
 \caption{Zero-shot judge selection prompt}
 \label{zs-judge-prompt}
\end{figure}

\begin{figure}[t!]
  \centering
  \includegraphics[width =1\linewidth]{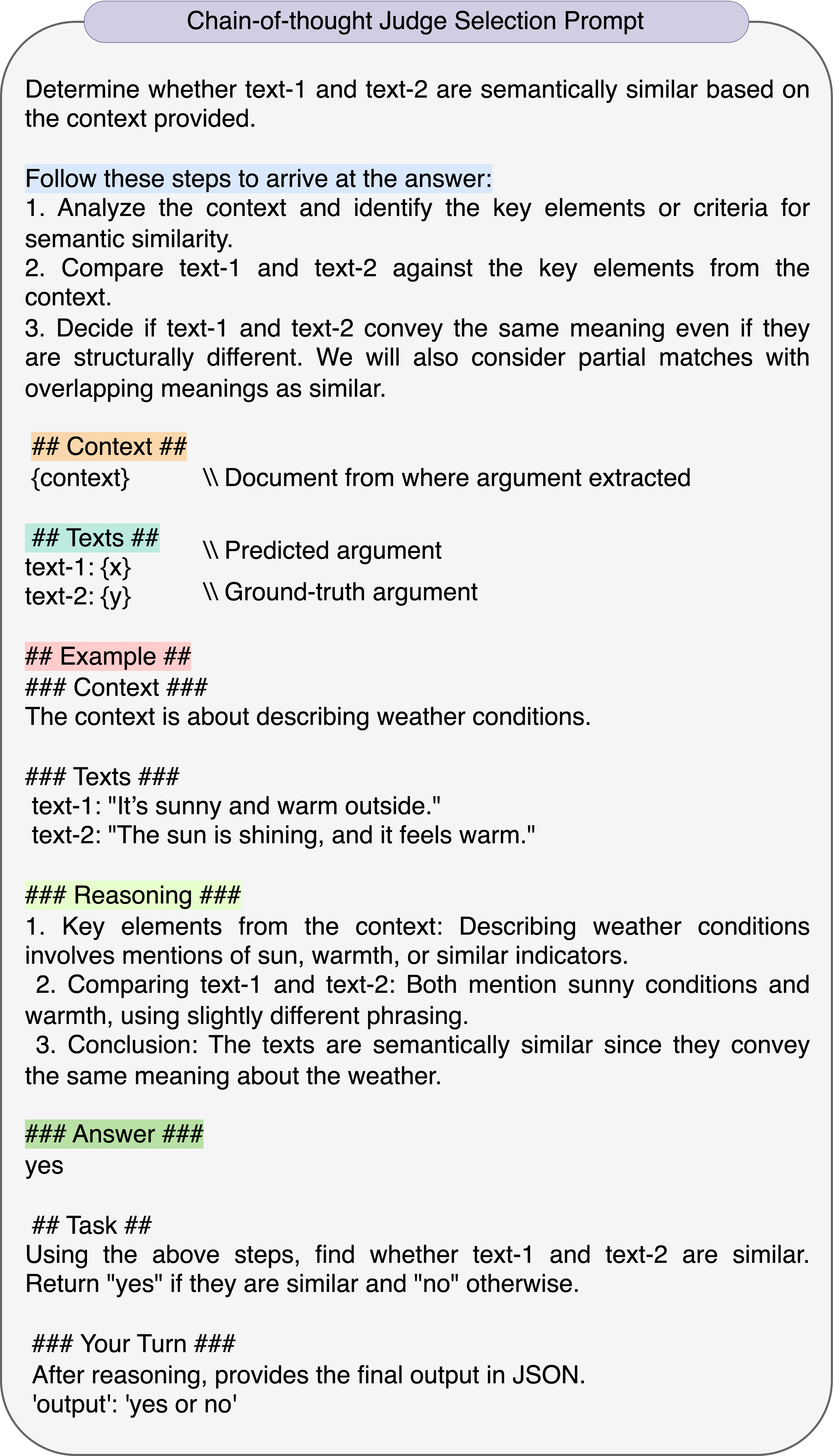}
 \caption{Chain-of-thought judge selection prompt}
 \label{cot-judge-prompt}
\end{figure}

First, we create a judge dataset through manual annotation. We then use this dataset to experiment with multiple models and select the most suitable judge. Specifically, we evaluate four models: Llama3.1-70B, GPT-3.5, GPT-4o-mini, and GPT-4o. Each model is tested using both \textit{zero-shot} and \textit{chain-of-thought} prompts, with a single prompt uniformly applied across all the evaluated datasets and models. Figures \ref{zs-judge-prompt} and \ref{cot-judge-prompt} show the prompts for zero-shot and chain-of-thought, respectively. Table \ref{LLM-Human-Aggrement-Rate} presents the agreement rate between the judge models and human evaluations. 

\begin{table}[h!]
\centering
\renewcommand*{\arraystretch}{1}
\small
\begin{tabular}{l|C{1cm}C{1.2cm}C{1.2cm}C{1.2cm}c}
& GPT-3.5 & GPT-4o-mini &  Llama3-70B & GPT-4o  \\
\toprule
ZS & 73.05 & 84.27 & 51.52 & \cellcolor{green!40} 86.17 \\
COT & 79.91 & 68.11 & 73.95 & 78.39\\
\bottomrule
\end{tabular}
\caption{Agreement percentage of different LLMs with human judgments. ZS and COT indicate zero-shot and chain-of-thought prompting approaches, respectively.} 
\label{LLM-Human-Aggrement-Rate}
\end{table}

While prompt optimization and alternative techniques (e.g., self-consistency) could further improve agreement, we refrain from such experiments due to the high cost and time requirements. Iterating to find optimal prompts for each model and dataset is impractical. These aspects, along with exploring the applicability of small fine-tuned judge models, are better suited for a separate study. 

\section{Additional Results}
\begin{table}[t!]
\centering
\renewcommand*{\arraystretch}{0.9}
\setlength{\tabcolsep}{4pt}
\small
\begin{tabular}{l|C{1cm}C{1cm}C{1cm}C{1.5cm}}
Dataset & Avg. EM-F1 & Avg. RM-F1 & Avg. CM-F1 & Avg. REGen-F1 (JAM-F1) \\
\midrule
DiscourseEE & 10.74 & 15.11 & 41.01 & 37.45 \\
PHEE        & 39.25 & 46.94 & 63.55 & 62.34 \\
RAMS        & 13.38 & 16.14 & 29.45 & 28.27 \\
GENEVA      & 13.62 & 25.32 & 48.18 & 46.12 \\
DocEE       & 17.33 & 24.98 & 45.12 & 41.65 \\
WikiEvents  &  8.93 & 11.05 & 33.56 & 31.02 \\
\bottomrule
\end{tabular}
\caption{Comparison of average F1-scores of LLMs under different evaluation frameworks: Exact Match (EM), Relaxed Match (RM), Complex Match (CM), and REGen.} 
\label{all-comparison-F1}
\end{table}

\subsection{Comparison with Head Noun Phrase Match Approach}
Following prior work \cite{du-cardie-2020-event, tong-etal-2022-docee}, we also evaluate model performance using the Head Noun Phrase Match (HM) approach for comprehensiveness. The results, shown in Table \ref{HM-improvement-rate}, indicate that on average models achieve 19\% higher F1 score with REGen than the HM approach across all datasets. This is consistent with the performance gain compared with the exact match approach (Table \ref{average-improvement-rate}).  We emphasize that, similar to exact and relaxed match strategies, the HM approach can result in inaccurate and misleading evaluations. This is because it only compares the head noun phrases in arguments, ignoring critical contextual information. For examples, for the role \texttt{date}, if the ground-truth is \texttt{`18 April 2024'} and predicted output is \texttt{`20 April 2018'}, the HM approach would consider them a match, as their noun phrase is `April', despite being semantically and factually different. Moreover, HM fails to assess arguments that do not contain noun phrases at all, resulting in unreliable evaluations.

\begin{table}[t!]
\centering
\renewcommand*{\arraystretch}{0.9}
\setlength{\tabcolsep}{4pt}
\small
\begin{tabular}{l|C{1cm}C{1.5cm}cC{1.2cm}}
Datasets& Avg. HM-F1 & Avg. REGen-F1 & $\Delta$ F1  &Gain (\%) \\
\toprule
DiscourseEE & 13.09 & 37.45 & +24.36 & +186.09 \\
PHEE       & 38.94 & 62.34 & +23.40 & +60.10 \\
RAMS       & 16.45 & 28.27 & +11.82 & +71.85 \\
GENEVA     & 25.50 & 46.12 & +20.62 & +80.86 \\
DocEE      & 22.20 & 41.65 & +19.45 & +87.61 \\
WikiEvents & 16.53  & 31.02 & +14.49 & +87.66 \\
\midrule
&\multicolumn{2}{c}{Avg. $\Delta$ F1}& +19.02 &\\
\bottomrule
\end{tabular}
\caption{Comparison of average F1-scores of the LLMs between Head Noun Phrase Match (HM) and REGen evaluation framework.} 
\label{HM-improvement-rate}
\end{table}

\begin{table*}[t!]
\centering
\renewcommand*{\arraystretch}{1}
\small
\begin{tabular}{l|cccccc|C{1.9cm}}
Datasets & DiscourseEE & PHEE & RAMS  &  GENEVA &  DocEE & WikiEvents & \\
\toprule
\multicolumn{7}{c}{Inference count and reduction in inference only for zero-shot approach}& Avg. Reduction (\%)  \\
\midrule
\#Inference (LLM as Judge) & 1822 & 6215 & 3718 & 3897 & 12655 & 1852 &  \\
\#Inference (REGen) & 1201 & 2077 & 2318 & 2465 & 6513 & 1158 & \\
Reduction count  & 621 & 4138 & 1400 & 1432 & 6142 & 694 & \\
\cmidrule{8-8}
Reduction (\%) & 34.08 & 66.58 & 37.65 & 36.74 & 48.53 & 37.47 & \textbf{43.51} \\
\midrule
\multicolumn{7}{c}{Inference count and reduction in inference only for only for chain-of-thought approach}&  \\
\midrule
\#Inference (LLM as Judge) & 1740 & 5991 & 3124 & 3549 & 11511 & 1588 &  \\
\#Inference (REGen) & 1186 & 2359 & 2200 & 2435 & 6111 & 1089 & \\
Reduction count  & 554 & 3632 & 924 & 1114 & 5400 & 499 & \\
\cmidrule{8-8}
Reduction (\%) & 31.83 & 60.62 & 29.57 & 31.38 & 46.91 & 31.42 & \textbf{38.62} \\
\midrule
\multicolumn{7}{c}{Total inference count and reduction in inference (zero-shot + chain-of-thought)}&  \\
\midrule
\#Inference (LLM as Judge) & 3562 & 12206 & 6842 & 7446 & 24166 & 3440 &  \\
\#Inference (REGen) & 2387 & 4436 & 4518 & 4900 & 12624 & 2247 & \\
Reduction count  & 1175 & 7770 & 2324 & 2546 & 11542 & 1193 & \\
\cmidrule{8-8}
Reduction (\%) & 32.98 & 63.65 & 33.96 & 34.19 & 47.76 & 34.68 & \textbf{41.20}\\
\bottomrule
\end{tabular}
\caption{Detailed comparison of inference counts and reductions when using LLMs as Judge versus the proposed REGen framework for the GPT-4o prediction model. Results for both zero-shot and chain-of-thought approaches are presented, illustrating total inference counts, achieved reductions, and corresponding percentage reductions across the evaluated datasets.} 
\label{inference-reduction-full-stat}
\end{table*}

%% DiscourseEE Results
\begin{table*}[h!]
\small
\centering
\renewcommand*{\arraystretch}{1}

\begin{tabular}{l|ccc|ccc|ccc|ccc}
 & \multicolumn{3}{c}{\textbf{Exact-Match}}& \multicolumn{3}{c}{\textbf{Relaxed-Match}} & \multicolumn{3}{c}{\textbf{Complex-Match}}& \multicolumn{3}{c}{\textbf{JAM-Score}} \\
\midrule
\textbf{Model}& P & R & F1 & P & R & F1  & P & R & F1 & P & R & F1\\
\midrule

\multicolumn{13}{c}{\textit{Baselines}}\\
\midrule
BERT & 6.3	& 5.52	& 5.88	& 9.28	& 8.12	& 8.66 &	35.17 & 32.1 & 33.56 & 31.65 & 28.84& 30.18 \\

Flan-T5 & 7.22 & 6.32 & 6.74 & 10.88 & 9.53 & 10.16 & 37.8 & 35.21 & 36.46 & 34.13 & 31.71 & 32.87 \\

\midrule
\multicolumn{13}{c}{\textit{LLMs with Zero-Shot Prompt}}\\
\midrule

\includegraphics[width=0.30cm, height=0.30cm]{icons/ms-icon.png} Phi-3.5 & 3.21 & 3.61 & 3.40 & 4.73 & 5.32 & 5.00 & 14.81 & 14.64 & 14.73 & 13.43 & 13.36 & 13.39 \\
\includegraphics[width=0.30cm, height=0.30cm]{icons/google-icon.png} Gemma-1.1 & 10.45 & 13.74 & 11.87 & 13.96 & 18.36 & 15.86 & 45.61 & 55.67 & 50.14 & 41.31 & 50.58 & 45.48 \\
\includegraphics[width=0.30cm, height=0.30cm]{icons/mistral-ai-icon.png} Mixtral & 10.59 & 17.15 & 13.10 & 14.37 & 23.17 & 17.74 & 41.82 & 57.97 & 48.59 & 38.07 & 53.19 & 44.38 \\
\includegraphics[width=0.30cm, height=0.30cm]{icons/meta-icon.png} Llama-3.1 & 11.05 & 16.95 & 13.38 & 15.49 & 23.67 & 18.73 & 37.52 & 51.96 & 43.57 & 34.47 & 48.02 & 40.13 \\
\includegraphics[width=0.32cm, height=0.32cm]{icons/openai-icon.png} GPT-4o & 14.14 & 20.76 & 16.82 & 19.40 & 28.49 & 23.08 & 43.99 & 57.57 & 49.87 & 40.58 & 53.50 & 46.16\\

\midrule
\multicolumn{13}{c}{\textit{LLMs with Chain-of-thought Prompt}}\\
\midrule

\includegraphics[width=0.30cm, height=0.30cm]{icons/ms-icon.png} Phi-3.5 & 5.53 & 9.83 & 7.08 & 9.76 & 17.05 & 12.42 & 34.26 & 52.36 & 41.41 & 30.89 & 47.47 & 37.43 \\
\includegraphics[width=0.30cm, height=0.30cm]{icons/google-icon.png} Gemma-1.1 & 7.60 & 12.14 & 9.35 & 10.62 & 16.95 & 13.06 & 37.44 & 51.25 & 43.27 & 33.79 & 46.57 & 39.16 \\
\includegraphics[width=0.30cm, height=0.30cm]{icons/mistral-ai-icon.png} Mixtral & 4.78 & 5.22 & 4.99 & 6.71 & 7.32 & 7.00 & 26.01 & 26.78 & 26.39 & 23.39 & 24.14 & 23.76 \\
\includegraphics[width=0.30cm, height=0.30cm]{icons/meta-icon.png} Llama-3.1 & 10.81 & 15.25 & 12.65 & 14.79 & 20.86 & 17.31 & 39.83 & 51.05 & 44.75 & 36.40 & 46.89 & 40.98 \\
\includegraphics[width=0.32cm, height=0.32cm]{icons/openai-icon.png} GPT-4o & 12.64 & 17.75 & 14.77 & 17.86 & 25.08 & 20.86 & 42.71 & 53.06 & 47.33 & 39.27 & 49.15 & 43.66 \\

\bottomrule 
\end{tabular}
\caption{DiscourseEE evaluation results using REGen framework.}
\label{DiscourseEE-all-results}
\end{table*}

%%PHEE dataset results
\begin{table*}[h!]
\small
\centering
\renewcommand*{\arraystretch}{1}

\begin{tabular}{l|ccc|ccc|ccc|ccc}
 & \multicolumn{3}{c}{\textbf{Exact-Match}}& \multicolumn{3}{c}{\textbf{Relaxed-Match}} & \multicolumn{3}{c}{\textbf{Complex-Match}}& \multicolumn{3}{c}{\textbf{JAM-Score}} \\
\midrule
\textbf{Model}& P & R & F1 & P & R & F1  & P & R & F1 & P & R & F1\\
\midrule

\multicolumn{13}{c}{\textit{Baselines }}\\
\midrule
BERT & 29.09 & 26.6 & 27.78 & 36.62 & 33.48 & 34.98 & 54.88 & 50.53 & 52.61 & 53.55 & 49.28 & 51.33 \\

Flan-T5 & 44.32 & 40.53 & 42.34 & 52.8 & 48.28 & 50.44 & 69.96 & 64.24 & 66.98 & 68.71 & 63.07 & 65.77 \\

\midrule
\multicolumn{13}{c}{\textit{LLMs with Zero-Shot Prompt}}\\
\midrule

\includegraphics[width=0.30cm, height=0.30cm]{icons/ms-icon.png} Phi-3.5 & 39.07 & 47.88 & 43.03 & 45.84 & 56.12 & 50.46 & 62.61 & 73.63 & 67.67 & 61.39 & 72.35 & 66.42\\
\includegraphics[width=0.30cm, height=0.30cm]{icons/google-icon.png} Gemma-1.1 & 44.10 & 45.94 & 45.00 & 53.23 & 55.49 & 54.34 & 75.91 & 77.99 & 76.93 & 74.25 & 76.35 & 75.28\\
\includegraphics[width=0.30cm, height=0.30cm]{icons/mistral-ai-icon.png} Mixtral & 33.68 & 40.02 & 36.58 & 39.16 & 46.59 & 42.55 & 55.39 & 63.55 & 59.19 & 54.20 & 62.31 & 57.98 \\
\includegraphics[width=0.30cm, height=0.30cm]{icons/meta-icon.png} Llama-3.1 & 36.49 & 42.29 & 39.17 & 43.77 & 50.63 & 46.95 & 61.53 & 66.60 & 63.96 & 60.23 & 65.43 & 62.72 \\
\includegraphics[width=0.32cm, height=0.32cm]{icons/openai-icon.png} GPT-4o & 51.38 & 56.18 & 53.67 & 59.34 & 64.74 & 61.92 & 77.36 & 80.63 & 78.96 & 76.04 & 79.47 & 77.72 \\

\midrule
\multicolumn{13}{c}{\textit{LLMs with Chain-of-thought Prompt}}\\
\midrule

\includegraphics[width=0.30cm, height=0.30cm]{icons/ms-icon.png} Phi-3.5 & 30.57 & 33.76 & 32.09 & 36.24 & 39.96 & 38.01 & 52.68 & 55.45 & 54.03 & 51.48 & 54.32 & 52.86 \\
\includegraphics[width=0.30cm, height=0.30cm]{icons/google-icon.png} Gemma-1.1 & 33.39 & 34.92 & 34.14 & 41.35 & 43.26 & 42.28 & 60.99 & 61.93 & 61.46 & 59.56 & 60.57 & 60.06 \\
\includegraphics[width=0.30cm, height=0.30cm]{icons/mistral-ai-icon.png} Mixtral & 29.12 & 29.81 & 29.46 & 36.87 & 37.70 & 37.28 & 50.90 & 50.61 & 50.75 & 49.87 & 49.66 & 49.77 \\
\includegraphics[width=0.30cm, height=0.30cm]{icons/meta-icon.png} Llama-3.1 & 30.44 & 32.19 & 31.29 & 38.87 & 41.05 & 39.93 & 52.20 & 52.83 & 52.51 & 51.22 & 51.97 & 51.59\\
\includegraphics[width=0.32cm, height=0.32cm]{icons/openai-icon.png} GPT-4o & 46.91 & 49.43 & 48.14 & 54.27 & 57.13 & 55.66 & 69.68 & 70.34 & 70.01 & 68.56 & 69.37 & 68.96 \\

\bottomrule 
\end{tabular}
\caption{PHEE evaluation results using REGen framework.}
\label{PHEE-all-results}
\end{table*}

%% RAMS Results
\begin{table*}[h!]
\small
\centering
\renewcommand*{\arraystretch}{1.1}

\begin{tabular}{l|ccc|ccc|ccc|ccc}
 & \multicolumn{3}{c}{\textbf{Exact-Match}}& \multicolumn{3}{c}{\textbf{Relaxed-Match}} & \multicolumn{3}{c}{\textbf{Complex-Match}}& \multicolumn{3}{c}{\textbf{JAM-Score}} \\
\midrule
\textbf{Model}& P & R & F1 & P & R & F1  & P & R & F1 & P & R & F1\\
\midrule

\multicolumn{13}{c}{\textit{Baselines }}\\
\midrule
BERT & 14.63 & 14.63 & 14.63 & 18.14 & 18.14 & 18.14 & 33.61 & 33.61 & 33.61 & 32.24 & 32.24 & 32.24\\
Flan-T5 & 12.61 & 12.61 & 12.61 & 15.13 & 15.13 & 15.13 & 28.62 & 28.62 & 28.62 & 27.43 & 27.43 & 27.43 \\

\midrule
\multicolumn{13}{c}{\textit{LLMs with Zero-Shot Prompt}}\\
\midrule

\includegraphics[width=0.30cm, height=0.30cm]{icons/ms-icon.png} Phi-3.5 & 13.75 & 17.35 & 15.34 & 16.07 & 20.27 & 17.92 & 31.23 & 37.77 & 34.19 & 29.90 & 36.22 & 32.76 \\
\includegraphics[width=0.30cm, height=0.30cm]{icons/google-icon.png} Gemma-1.1 & 14.40 & 15.37 & 14.87 & 16.95 & 18.09 & 17.50 & 31.45 & 33.47 & 32.43 & 30.17 & 32.11 & 31.11 \\
\includegraphics[width=0.30cm, height=0.30cm]{icons/mistral-ai-icon.png} Mixtral & 11.30 & 15.22 & 12.97 & 13.47 & 18.14 & 15.46 & 26.42 & 34.50 & 29.93 & 25.28 & 33.06 & 28.65 \\
\includegraphics[width=0.30cm, height=0.30cm]{icons/meta-icon.png} Llama-3.1 & 9.98 & 14.88 & 11.95 & 12.20 & 18.04 & 14.56 & 21.82 & 30.50 & 25.44 &  20.96 & 29.39 & 24.47 \\
\includegraphics[width=0.32cm, height=0.32cm]{icons/openai-icon.png} GPT-4o & 15.01 & 27.58 & 19.44 & 17.89 & 32.82 & 23.15 & 29.91 & 49.98 & 37.42 & 28.84 & 48.43 & 36.15 \\

\midrule
\multicolumn{13}{c}{\textit{LLMs with Chain-of-thought Prompt}}\\
\midrule

\includegraphics[width=0.30cm, height=0.30cm]{icons/ms-icon.png} Phi-3.5 & 14.15 & 17.20 & 15.53 & 17.00 & 20.66 & 18.65 & 31.96 & 37.57 & 34.54 & 30.64 & 36.07 & 33.13\\
\includegraphics[width=0.30cm, height=0.30cm]{icons/google-icon.png} Gemma-1.1 & 10.17 & 11.32 & 10.71 & 12.88 & 14.34 & 13.57 & 25.13 & 27.53 & 26.28 & 24.04 & 26.36 & 25.15 \\
\includegraphics[width=0.30cm, height=0.30cm]{icons/mistral-ai-icon.png} Mixtral & 6.08 & 7.86 & 6.85 & 7.34 & 9.49 & 8.28 & 15.33 & 19.08 & 17.00 & 14.63 & 18.23 & 16.23 \\
\includegraphics[width=0.30cm, height=0.30cm]{icons/meta-icon.png} Llama-3.1 & 9.28 & 12.51 & 10.66 & 11.12 & 14.98 & 12.76 & 21.21 & 26.89 & 23.72 & 20.32 & 25.83 & 22.75 \\
\includegraphics[width=0.32cm, height=0.32cm]{icons/openai-icon.png} GPT-4o & 12.77 & 19.72 & 15.50 & 16.13 & 24.91 & 19.58 & 28.55 & 40.68 & 33.56 & 27.44 & 39.26 & 32.30 \\

\bottomrule 
\end{tabular}
\caption{RAMS evaluation results using REGen framework.}
\label{RAMS-all-results}
\end{table*}

%GENEVA Results
\begin{table*}[h!]
\small
\centering
\renewcommand*{\arraystretch}{1.1}

\begin{tabular}{l|ccc|ccc|ccc|ccc}
 & \multicolumn{3}{c}{\textbf{Exact-Match}}& \multicolumn{3}{c}{\textbf{Relaxed-Match}} & \multicolumn{3}{c}{\textbf{Complex-Match}}& \multicolumn{3}{c}{\textbf{JAM-Score}} \\
\midrule
\textbf{Model}& P & R & F1 & P & R & F1  & P & R & F1 & P & R & F1\\
\midrule

\multicolumn{13}{c}{\textit{Baselines}}\\
\midrule
BERT & 15.81 & 14.72 & 15.24 & 27.56 & 25.67 & 26.58 & 54.64 & 51.62 & 53.09 & 52.24 & 49.33 & 50.74\\

Flan-T5 & 19.02 & 17.71 & 18.34 & 32.0 & 29.79 & 30.85 & 59.63 & 56.01 & 57.76 & 57.16 & 53.67 & 55.36\\

\midrule
\multicolumn{13}{c}{\textit{LLMs with Zero-Shot Prompt}}\\
\midrule

\includegraphics[width=0.30cm, height=0.30cm]{icons/ms-icon.png} Phi-3.5 & 12.21 & 14.36 & 13.20 & 23.56 & 27.68 & 25.46 & 47.65 & 52.14 & 49.80 & 45.50 & 49.92 & 47.61 \\
\includegraphics[width=0.30cm, height=0.30cm]{icons/google-icon.png} Gemma-1.1 & 11.75 & 11.63 & 11.69 & 24.52 & 24.27 & 24.40 & 51.41 & 50.06 & 50.73 & 49.01 & 47.75 & 48.37 \\
\includegraphics[width=0.30cm, height=0.30cm]{icons/mistral-ai-icon.png} Mixtral & 12.20 & 14.65 & 13.31 & 22.80 & 27.32 & 24.86 & 45.98 & 51.17 & 48.44 & 43.92 & 49.01 & 46.32 \\
\includegraphics[width=0.30cm, height=0.30cm]{icons/meta-icon.png} Llama-3.1 & 15.04 & 17.93 & 16.36 & 27.24 & 32.46 & 29.62 & 52.74 & 57.67 & 55.09 & 50.45 & 55.36 & 52.79 \\
\includegraphics[width=0.32cm, height=0.32cm]{icons/openai-icon.png} GPT-4o & 17.72 & 20.86 & 19.16 & 30.86 & 36.29 & 33.35 & 56.25 & 60.49 & 58.30 & 53.96 & 58.25 & 56.02 \\

\midrule
\multicolumn{13}{c}{\textit{LLMs with Chain-of-thought Prompt}}\\
\midrule

\includegraphics[width=0.30cm, height=0.30cm]{icons/ms-icon.png} Phi-3.5 & 9.98 & 11.66 & 10.76 & 20.22 & 23.55 & 21.76 & 44.22 & 48.60 & 46.31 & 42.09 & 46.36 & 44.12 \\
\includegraphics[width=0.30cm, height=0.30cm]{icons/google-icon.png} Gemma-1.1 & 9.21 & 9.55 & 9.38 & 20.83 & 21.60 & 21.21 & 45.69 & 45.32 & 45.51 & 43.47 & 43.18 & 43.33 \\
\includegraphics[width=0.30cm, height=0.30cm]{icons/mistral-ai-icon.png} Mixtral & 15.39 & 17.48 & 16.37 & 25.17 & 28.59 & 26.77 & 45.08 & 49.25 & 47.07 & 43.29 & 47.38 & 45.24 \\
\includegraphics[width=0.30cm, height=0.30cm]{icons/meta-icon.png} Llama-3.1 & 9.29 & 9.65 & 9.46 & 16.98 & 17.61 & 17.29 & 33.15 & 31.03 & 32.05 & 31.70 & 29.79 & 30.72\\
\includegraphics[width=0.32cm, height=0.32cm]{icons/openai-icon.png} GPT-4o & 16.00 & 17.12 & 16.54 & 27.57 & 29.50 & 28.50 & 48.50 & 48.47 & 48.48 & 46.59 & 46.71 & 46.65 \\

\bottomrule 
\end{tabular}
\caption{GENEVA evaluation results using REGen framework.}
\label{GENEVA-all-results}
\end{table*}

%DocEE Results
\begin{table*}[h!]
\small
\centering
\renewcommand*{\arraystretch}{1}
\begin{tabular}{l|ccc|ccc|ccc|ccc}
 & \multicolumn{3}{c}{\textbf{Exact-Match}}& \multicolumn{3}{c}{\textbf{Relaxed-Match}} & \multicolumn{3}{c}{\textbf{Complex-Match}}& \multicolumn{3}{c}{\textbf{JAM-Score}} \\
\midrule
\textbf{Model}& P & R & F1 & P & R & F1  & P & R & F1 & P & R & F1\\
\midrule

\multicolumn{13}{c}{\textit{Baselines }}\\
\midrule
BERT & 22.93 & 15.73 & 18.66 & 31.59 & 21.72 & 25.74 & 57.09 & 41.12 & 47.81 & 52.73 & 37.82 & 44.05 \\

Flan-T5 & 22.8 & 15.64 & 18.55 & 30.62 & 21.08 & 24.97 & 54.35 & 38.98 & 45.4 & 50.29 & 35.94 & 41.92 \\

\midrule
\multicolumn{13}{c}{\textit{LLMs with Zero-Shot Prompt}}\\
\midrule

\includegraphics[width=0.30cm, height=0.30cm]{icons/ms-icon.png} Phi-3.5 & 14.16 & 14.36 & 14.26 & 19.86 & 20.04 & 19.95 & 39.18 & 37.62 & 38.39 & 35.90 & 34.62 & 35.25 \\
\includegraphics[width=0.30cm, height=0.30cm]{icons/google-icon.png} Gemma-1.1 & 20.47 & 16.04 & 17.99 & 30.38 & 23.95 & 26.78 & 51.92 & 42.54 & 46.77 & 48.15 & 39.31 & 43.28 \\
\includegraphics[width=0.30cm, height=0.30cm]{icons/mistral-ai-icon.png} Mixtral & 21.84 & 24.09 & 22.91 & 31.06 & 34.20 & 32.55 & 56.50 & 59.92 & 58.16 & 52.12 & 55.47 & 53.74 \\
\includegraphics[width=0.30cm, height=0.30cm]{icons/meta-icon.png} Llama-3.1 & 15.03 & 21.11 & 17.56 & 21.65 & 29.97 & 25.14 & 42.51 & 51.17 & 46.44 & 38.95 & 47.49 & 42.80 \\
\includegraphics[width=0.32cm, height=0.32cm]{icons/openai-icon.png} GPT-4o & 16.82 & 31.42 & 21.91 & 24.45 & 44.89 & 31.65 & 45.79 & 73.41 & 56.40 & 42.12 & 68.41 & 52.14 \\

\midrule
\multicolumn{13}{c}{\textit{LLMs with Chain-of-thought Prompt}}\\
\midrule

\includegraphics[width=0.30cm, height=0.30cm]{icons/ms-icon.png} Phi-3.5 & 18.95 & 20.82 & 19.84 & 26.57 & 29.13 & 27.79 & 46.94 & 50.19 & 48.51 & 43.43 & 46.55 & 44.93 \\
\includegraphics[width=0.30cm, height=0.30cm]{icons/google-icon.png} Gemma-1.1 & 10.90 & 9.01 & 9.87 & 16.20 & 13.47 & 14.71 & 28.44 & 23.98 & 26.02 & 26.30 & 22.15 & 24.05 \\
\includegraphics[width=0.30cm, height=0.30cm]{icons/mistral-ai-icon.png} Mixtral & 7.84 & 6.17 & 6.90 & 10.93 & 8.60 & 9.63 & 22.38 & 16.94 & 19.28 & 20.44 & 15.53 & 17.65 \\
\includegraphics[width=0.30cm, height=0.30cm]{icons/meta-icon.png} Llama-3.1 & 16.81 & 24.07 & 19.79 & 25.26 & 35.97 & 29.68 & 48.02 & 61.66 & 53.99 & 44.10 & 57.15  & 49.78 \\
\includegraphics[width=0.32cm, height=0.32cm]{icons/openai-icon.png} GPT-4o & 17.78 & 29.77 & 22.26 & 25.56 & 42.48 & 31.92 & 47.90 & 71.18 & 57.27 & 44.07 & 66.17 & 52.90\\

\bottomrule 
\end{tabular}
\caption{DocEE evaluation results using REGen framework.}
\label{DocEE-all-results}
\end{table*}

%WikiEvents Results
\begin{table*}[h!]
\small
\centering
\renewcommand*{\arraystretch}{1}

\begin{tabular}{l|ccc|ccc|ccc|ccc}
 & \multicolumn{3}{c}{\textbf{Exact-Match}}& \multicolumn{3}{c}{\textbf{Relaxed-Match}} & \multicolumn{3}{c}{\textbf{Complex-Match}}& \multicolumn{3}{c}{\textbf{JAM-Score}} \\
\midrule
\textbf{Model}& P & R & F1 & P & R & F1  & P & R & F1 & P & R & F1\\
\midrule

\multicolumn{13}{c}{\textit{Baselines}}\\
\midrule
BERT & 9.62 & 4.86 & 6.46 & 14.23 & 7.19 & 9.55 & 38.91 & 23.68 & 29.44 & 36.12 & 21.82 & 27.2 \\

Flan-T5 & 13.81 & 6.98 & 9.27 & 17.57 & 8.88 & 11.8 & 39.33 & 23.47 & 29.4 & 36.87 & 21.82 & 27.41 \\

\midrule
\multicolumn{13}{c}{\textit{LLMs with Zero-Shot Prompt}}\\
\midrule

\includegraphics[width=0.30cm, height=0.30cm]{icons/ms-icon.png} Phi-3.5 & 9.80 & 8.46 & 9.08 & 11.76 & 10.15 & 10.90 & 36.76 & 32.56 & 34.53 & 33.94 & 30.03 & 31.86 \\
\includegraphics[width=0.30cm, height=0.30cm]{icons/google-icon.png} Gemma-1.1 & 8.65 & 4.86 & 6.22 & 10.15 & 5.71 & 7.31 & 40.98 & 29.60 & 34.37 & 37.49 & 26.90 & 31.32 \\
\includegraphics[width=0.30cm, height=0.30cm]{icons/mistral-ai-icon.png} Mixtral & 10.55 & 9.30 & 9.89 & 13.19 & 11.63 & 12.36 & 39.09 & 37.42 & 38.24 & 36.16 & 34.51 & 35.31 \\
\includegraphics[width=0.30cm, height=0.30cm]{icons/meta-icon.png} Llama-3.1 & 11.06 & 15.22 & 12.81 & 13.36 & 18.39 & 15.48 & 33.95 & 45.67 & 38.94 & 31.62 & 42.58 & 36.29 \\
\includegraphics[width=0.32cm, height=0.32cm]{icons/openai-icon.png} GPT-4o & 11.47 & 17.34 & 13.80 & 14.13 & 21.35 & 17.00 & 35.10 & 51.80 & 41.85 & 32.73 & 48.36 & 39.04 \\

\midrule
\multicolumn{13}{c}{\textit{LLMs with Chain-of-thought Prompt}}\\
\midrule

\includegraphics[width=0.30cm, height=0.30cm]{icons/ms-icon.png} Phi-3.5 & 7.82 & 6.13 & 6.87 & 9.70 & 7.61 & 8.53 & 35.04 & 29.39 & 31.97 & 32.18 & 26.93 & 29.32 \\
\includegraphics[width=0.30cm, height=0.30cm]{icons/google-icon.png} Gemma-1.1 & 4.53 & 2.54 & 3.25 & 6.42 & 3.59 & 4.61 & 22.26 & 14.59 & 17.63 & 20.47 & 13.35 & 16.16 \\
\includegraphics[width=0.30cm, height=0.30cm]{icons/mistral-ai-icon.png} Mixtral & 5.96 & 3.59 & 4.49 & 8.07 & 4.86 & 6.07 & 24.56 & 16.28 & 19.58 & 22.70  & 14.99 & 18.06 \\
\includegraphics[width=0.30cm, height=0.30cm]{icons/meta-icon.png} Llama-3.1 & 10.78 & 10.78 & 10.78 & 13.11 & 13.11 & 13.11 & 35.94 & 37.21 & 36.56 & 33.36 & 34.49 & 33.91\\
\includegraphics[width=0.32cm, height=0.32cm]{icons/openai-icon.png} GPT-4o & 10.77 & 13.95 & 12.15 & 13.38 & 17.34 & 15.10 & 37.36 & 47.78 & 41.93 & 34.65 & 44.34 & 38.90 \\

\bottomrule 
\end{tabular}
\caption{WikiEvents evaluation results using REGen framework.}
\label{WikiEvents-all-results}
\end{table*}

\begin{figure}[t!]
  \centering
  \includegraphics[width =1\linewidth]{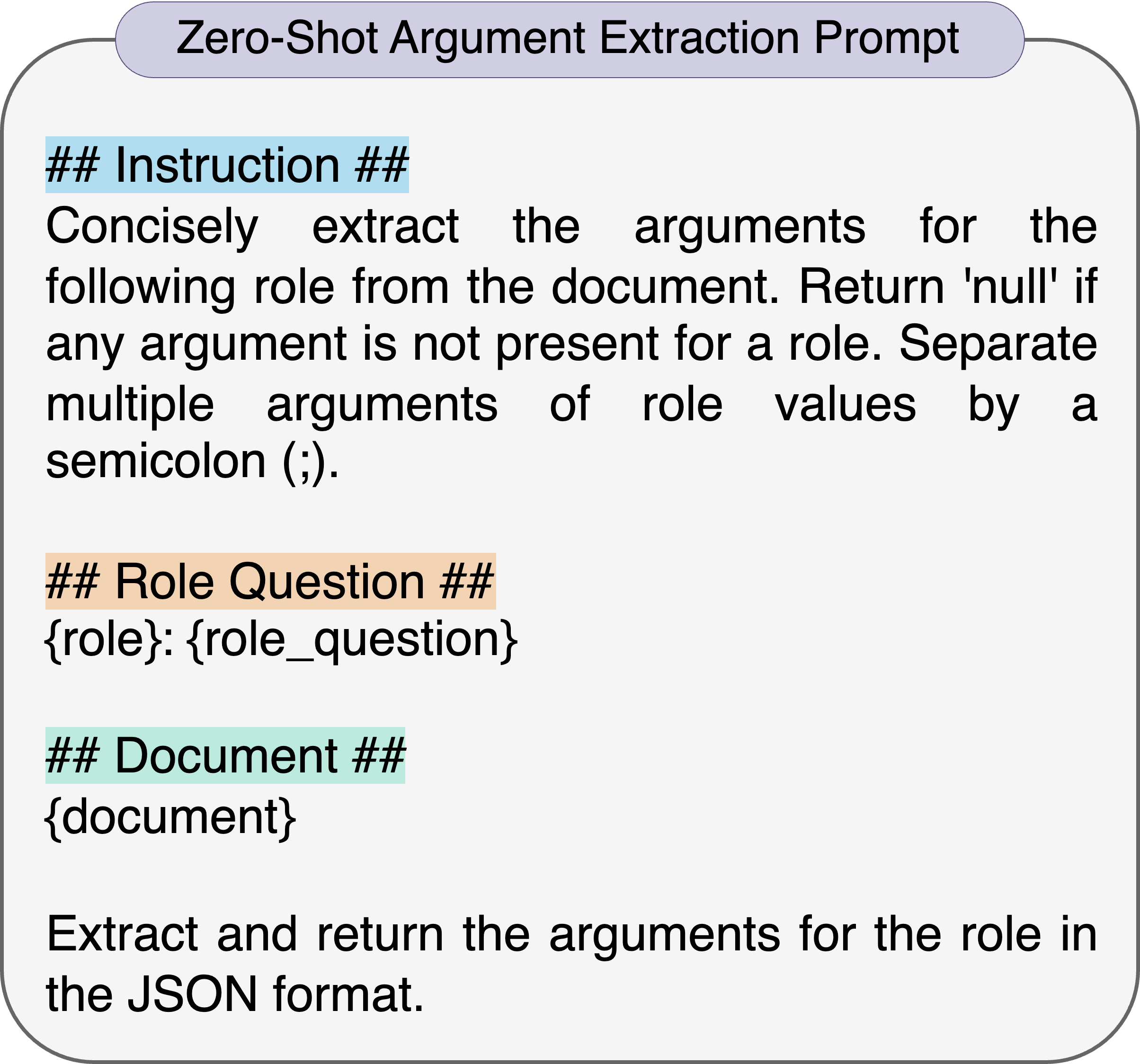}
 \caption{Zero-shot event argument extraction prompt}
 \label{zs-EAE-prompt}
\end{figure}

\begin{figure}[t!]
  \centering
  \includegraphics[width =1\linewidth]{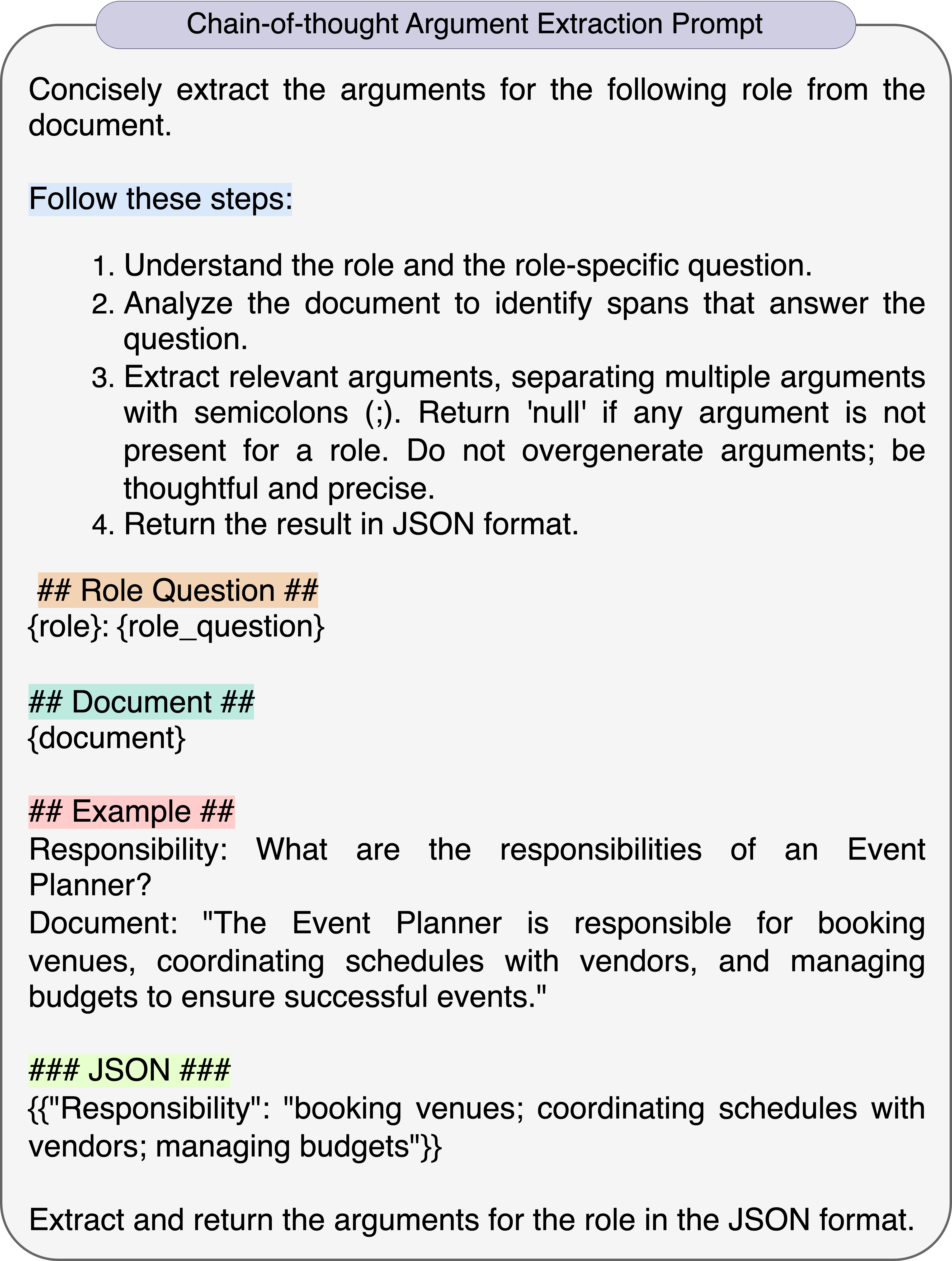}
 \caption{Chain-of-thought event argument extraction prompt}
 \label{cot-EAE-prompt}
\end{figure}

\begin{table*}[h!]
\centering
\renewcommand*{\arraystretch}{1}
\small
\begin{tabular}{cL{4.2cm}L{2.4cm}L{6cm}}

\toprule
\textbf{Dataset}&\textbf{Event} & \textbf{Role} & \textbf{Question}\\
\midrule

\multirow{10}{*}{\texttt{DiscourseEE}} & \texttt{Taking-MOUD} & \texttt{Treatment} & What treatments the subject/patient prescribed or undergoing? \\ 
&\texttt{Tapering} & \texttt{Side effects} & What are the side effects the subject is experiencing or expects to experience? \\ 
& \texttt{Return to Usage} & \texttt{Intervention} & What measures are taken to address or reduce side effects?\\ 
& \texttt{Taking-MOUD} & \texttt{Dosage} & What is the current or previous dosage of the Medications? \\ 
&\texttt{Tapering} & \texttt{Age} & What is the age of the subject/patient? \\ 
& \texttt{Return to Usage} & \texttt{Conditions} & What are the Pre-existing or co-morbid conditions of the subject/patient?\\ 
\midrule

\multirow{10}{*}{\texttt{PHEE}} & \texttt{Potential therapeutic event} & \texttt{Treatment} & What is the therapy administered to the patients? \\ 
& \texttt{Adverse event} & \texttt{Treatment drug} & Whare the the drugs used as therapy in the event? \\ 
& \texttt{Potential therapeutic event} & \texttt{Treatment dosage} & What is the amount of drug is given? \\ 
& \texttt{Adverse event} & \texttt{Treatment route} & What is the route of drug administration? \\ 
& \texttt{Adverse event} & \texttt{Effect} & What are the outcomes or side effects of the treatments? \\ 
& \texttt{Potential therapeutic event} & \texttt{Treatment disorder} & What is the target disorder of the medicine administration? \\ 
\midrule

\multirow{7}{*}{\texttt{RAMS}} & \texttt{Artifactexistence} & \texttt{Place} & Where does this event occur? \\ 
& \texttt{Transaction } & \texttt{Artifact} & What artifact is involved? \\ 
& \texttt{Contact.commandorder} & \texttt{Communicator} & Who is the communicator? \\ 
& \texttt{Movement transportartifact} & \texttt{Origin} & Where does the movement originate? \\ 
& \texttt{Conflict.yield.retreat} & \texttt{Retreater} & Who is the retreater? \\ 
& \texttt{transaction transferownership } & \texttt{Recipient} & Who is the recipient? \\ 
\midrule

\multirow{6}{*}{\texttt{GENEVA}} & \texttt{Statement} & \texttt{Message} & What is the message? \\
& \texttt{Collaboration} & \texttt{Partners} & Who are the partners in this collaboration? \\
& \texttt{Supply} & \texttt{Supplier} & Who is the supplier? \\
& \texttt{Protest} & \texttt{Content} & What is the content of the protest?\\
& \texttt{Killing} & \texttt{Victim} & Who is the victim?\\
& \texttt{Research} & \texttt{Topic} &What is the research topic?\\
\midrule

\multirow{6}{*}{\texttt{DocEE}} & \texttt{Riot} & \texttt{Location} & Where did the riot occur? \\
& \texttt{Regime change} & \texttt{Date} & When did the change happen? \\
& \texttt{Earthquakes} & \texttt{Affected area} & Which area was affected by the earthquake? \\
& \texttt{Military exercise} & \texttt{Scale} & What was the scale of the exercise? \\
& \texttt{Diplomatic talks} & \texttt{Participants} & Who are the participants? \\
& \texttt{Fire} & \texttt{Location} & Where did the fire take place? \\
\midrule
\multirow{6}{*}{\texttt{WikiEvents}} & \texttt{Conflict.attack} & \texttt{Instrument} & What instrument is used? \\
& \texttt{Life.die} & \texttt{Place} & Where did the death occur? \\
& \texttt{Conflict.detonateexplode} & \texttt{Target} & Who or what is the target? \\
& \texttt{Movement.transportation} & \texttt{Transporter} & Who is the transporter? \\
& \texttt{Justice.chargeindict} & \texttt{Defendant} & Who is the defendant? \\
& \texttt{Transaction} & \texttt{Acquired entity} & What entity is being acquired? \\
\bottomrule
\end{tabular}
 
\captionof{table}{\label{role-specific-question-details}Details of the argument roles for each event type in the evaluated datasets. Note: for RAMS and WikiEvents datasets some event names are very long. For presentation convenience we use the first part of the event name. }
\end{table*}

\end{document}